\documentclass[acmsmall]{acmart}

\usepackage{float}
\usepackage{xr}

\usepackage{listings}
\usepackage{graphicx}
\usepackage{textcomp}
\usepackage{xcolor}
\usepackage{subfig}
\usepackage{hyperref}
\usepackage[linesnumbered,ruled]{algorithm2e}
\usepackage{algorithmic}
\usepackage{multirow}
\usepackage{xfrac}
\usepackage{inputenc}
\usepackage{booktabs}
\usepackage{makecell}
\usepackage{graphicx}
\usepackage{enumitem}
\usepackage{diagbox}

\newcommand{\TheName}{\textbf{P$^2$ANet}}
\newcommand{\etal}{\textit{et al.}}

\AtBeginDocument{%
  \providecommand\BibTeX{{%
    \normalfont B\kern-0.5em{\scshape i\kern-0.25em b}\kern-0.8em\TeX}}}

\setcopyright{acmcopyright}
\copyrightyear{2018}
\acmYear{2018}
\acmDOI{XXXXXXX.XXXXXXX}

\acmJournal{JACM}
\acmVolume{37}
\acmNumber{4}
\acmArticle{111}
\acmMonth{8}

\begin{document}

\title{\TheName: A Large-Scale Benchmark for Dense Action Detection from Table Tennis Match Broadcasting Videos}

\author{Jiang Bian}
\email{jiangbian03@gmail.com}
\author{Xuhong Li}
\email{lixuhong@baidu.com}
\author{Tao Wang}
\email{wangtao@bupt.edu.cn}
\author{Qingzhong Wang}
\email{qingzwang@outlook.com}
\author{Jun Huang}
\email{huangjun12@baidu.com}
\author{Chen Liu}
\email{ttjygbtj@gmail.com}
\author{Jun Zhao}
\email{zhaojun12@baidu.com}
\author{Feixiang Lu}
\email{lufeixiang@baidu.com}
\author{Dejing Dou}
\email{doudejing@baidu.com}
\author{Haoyi Xiong}
\authornote{The corresponding author is Haoyi Xiong.}
\email{haoyi.xiong.fr@ieee.org}

\affiliation{%
  \institution{Baidu Inc.}
  \streetaddress{10 Shangdi 10th St}
  \city{Haidian District}
  \state{Beijing}
  \postcode{100094}
  \country{China}
}


\renewcommand{\shortauthors}{Jiang Bian, et al.}

\begin{abstract}
  While deep learning has been widely used for video analytics, such as video classification and action detection, dense action detection with fast-moving subjects from sports videos is still challenging. In this work, we release yet another sports video benchmark \TheName{} for \emph{\underline{P}}ing \emph{\underline{P}}ong-\emph{\underline{A}}ction detection, which consists of 2,721 video clips collected from the broadcasting videos of professional table tennis matches in World Table Tennis Championships and Olympiads. We work with a crew of table tennis professionals and referees on a specially designed annotation toolbox to obtain fine-grained action labels (in 14 classes) for every ping-pong action that appeared in the dataset, and formulate two sets of action detection problems---\emph{action localization} and \emph{action recognition}. We evaluate a number of commonly-seen action recognition (e.g., TSM, TSN, Video SwinTransformer, and Slowfast) and action localization models (e.g., BSN, BSN++, BMN, TCANet), using \TheName{} for both problems, under various settings. These models can only achieve 48\% area under the AR-AN curve for localization and 82\% top-one accuracy for recognition since the ping-pong actions are dense with fast-moving subjects but broadcasting videos are with only 25 FPS. The results confirm that \TheName{} is still a challenging task and can be used as a special benchmark for dense action detection from videos.
\end{abstract}





\maketitle

\section{Introduction}\label{sec:into}

Videos have become one of the most popular media in our everyday life and video understanding draws much attention from researchers in recent years, including video tagging \cite{yue2015beyond,abu2016youtube,wang2019knowledge}, retrieval \cite{gabeur2020multi,liu2019use,dong2021dual}, action recognition \cite{karpathy2014large,feichtenhofer2019slowfast,wei2021masked} and localization \cite{nawhal2021activity,xu2021boundary,dai2017temporal}. Video understanding has many applications, one of them being recommendation systems \cite{cai2022heterogeneous,yan2019multi,liu2019building}. One of the most attractive applications is understanding sports videos, which is able to benefit coaching \cite{wadsworth2020use,bao2021dynamic},  player training \cite{bertasius2017baller,sri2021toward} and sports broadcasting \cite{liu2017deep,wang2019football}. To better understand sports videos, localizing and recognizing the actions in untrimmed videos play crucial roles, yet they present challenging tasks and unsolved problems.

First, there are many types of sports, such as team sports like football, basketball, volleyball, and individual sports like tennis, table tennis, golf, and gymnastics, and each of them has specific actions. Therefore, it is difficult to build a dataset covering all sports and their specific actions. Moreover, the duration of actions in different sports is diverse. For example, the strokes in tennis and table tennis are extremely fast, usually less than one second, while actions in soccer could endure for several seconds, such as long passes and corner balls. Hence, it is difficult to model the actions with various lengths using one model. 

Second, data annotation of sports actions is challenging. Compared with common actions in our daily life like walking, and riding bicycles that can be easily recognized by annotators, it could be difficult to identify the actions in sports videos, for example, Axel jump, Flip jump, and Lutz jump in figure skating, hence, professional players should be involved in data annotation. In addition, players normally perform deceptive actions, for example, table tennis players can perform similar actions, but serve balls with different kinds of spin (top spin, back spin, and side spin), which is difficult for annotators to distinguish one action from the others.

Recently, researchers pay much attention to sports videos to address the challenges, including building datasets \cite{shao2020finegym,jiang2020soccerdb,martin2018sport,kulkarni2021table,deliege2021soccernet} and proposing new models \cite{martin2021three,koshkina2021contrastive,thilakarathne2021pose,bian2022machine}. In this paper, we focus on a specific sport -- table tennis and propose a benchmark -- \textit{Ping Pang Action} (\TheName{}) for action recognition and localization to facilitate research on fine-grained action understanding. \TheName{} is a unique framework that enables intelligent data acquisition via a specially designed annotation toolbox, and professional calibration via incorporating expert-in-domain refinement, thus, achieving a large-scale (the number of samples and the variety of categories), comprehensive (the fast-moving and dense characteristics), and easy-to-use video action dataset, named as \TheName{}. The key properties of \TheName{} are as follows: 
\begin{itemize}[leftmargin=*]
    \item We annotate each stroke in videos, including the category of the stroke and the indices of the starting and end frames. Plus, the stroke labels are confirmed by professional players, including Olympic table tennis players.
    \item To the best of our knowledge, \TheName{} is the largest dataset for table tennis analysis, composed of 2,721 untrimmed broadcasting videos, and the total length is 272 hours. Though \cite{martin2018sport,kulkarni2021table} propose table tennis datasets, the number of videos is smaller and they use self-recorded videos where there is only one player, which is much easier than using broadcasting videos. Moreover, the datasets proposed in \cite{martin2018sport,kulkarni2021table} are only for action recognition, whereas \TheName{} can also be used for action localization.
    \item The actions are \textbf{fast} and \textbf{dense} which is unique compared to existing sports datasets~\cite{ibrahim2016hierarchical,zalluhoglu2020collective,niebles2010modeling,soomro2014action,caba2015activitynet,faulkner2017tenniset,martin2018sport,voeikov2020ttnet,shao2020finegym,wu2022survey}. The action length ranges from 0.3 seconds to 3 seconds, but more than 90\% of actions are less than 1 second, which is around 10 times shorter than the average action duration in the existing sports datasets. In addition, there are around 15 actions in 10 seconds for \TheName{}, which is also an extremely high density (about 5$\times$) compared to the regular sports datasets, leading to a challenge for fine-grained localization. 
\end{itemize} 
With the properties, \TheName{} is a rich benchmark for research on action recognition and localization, especially for those \textbf{fast-moving} and \textbf{dense} actions. As shown in Table~\ref{tab:comparison}, the density and action length differ a lot from the existing representative sports datasets. \TheName{} further provides several existing widely used action recognition and localization models using \TheName{}, finding that \TheName{} is relatively challenging for both recognition and localization.

To sum up, the main contributions of this work are twofold. First, we develop a new challenging dataset -- \TheName{} for \textbf{fast} action recognition and \textbf{dense} action localization, which provides high-quality annotations confirmed by professional table tennis players. Compared with existing table tennis datasets, \TheName{} uses broadcasting videos instead of self-recorded ones, making it more challenging and flexible. In addition, \TheName{} is the largest one for table tennis analysis. Second, we benchmark a number of existing recognition and localization models using \TheName{}, finding that it is difficult to localize dense actions and recognize actions with imbalance categories, which could inspire researchers to come up with novel models for \textbf{dense} and \textbf{fast} action detection in the future.

\section{Related Work}

\subsection{Action Localization}

The task of action localization is to find the beginning and the end frames of actions in untrimmed videos. Normally, there are three steps in an action localization model. The first step is video feature extraction using deep neural networks. Second, classification, and finally, suppressing the noisy predictions. Z. Shou \etal \cite{shou2016temporal} propose a temporal action localization model based on multi-stage CNNs, where a deep proposal network is employed to identify the candidate segments of a long untrimmed video that contains actions, then a classification network is applied to predict the action labels of the candidate segments and finally, the localization network is fine-tuned. Alternatively, J. Yuan \etal \cite{yuan2016temporal} employ \textit{Improved Dense Trajectories} (IDT) to localize actions, however, IDT is time-consuming and requires more memory. In contrast, S. Yeung \etal \cite{yeung2016end} treat action localization as a decision-making process, where an agent observes video frames and decides where to look next and when to make a prediction. Similar to Faster-RCNN \cite{ren2015faster}, X. Dai \etal \cite{dai2017temporal} propose a temporal context network (TCN) for active localization, where proposal anchors are generated and the proposals with high scores are fed into the classification network to obtain the action categories. To generate better temporal proposals, T. Lin \etal \cite{lin2018bsn} propose a boundary-sensitive network (BSN), which adopts a local-to-global fashion, however, BSN is not a unified framework for action localization and the proposal feature is too simple to capture temporal context. To address the issues of BSN, T. Lin \etal \cite{lin2019bmn} propose a boundary matching network (BMN). Recently, M. Xu \etal \cite{xu2021boundary} introduce the pre-training-fine-tuning paradigm into action localization, where the model is first pre-trained on a boundary-sensitive synthetic dataset and then fine-tuned using the human-annotated untrimmed videos. Instead of using CNNs to extract video features, M. Nawhal \etal \cite{nawhal2021activity} employs a graph transformer, which significantly improves the performance of temporal action localization. 

\subsection{Action Recognition}
Action recognition lies at the heart of video understanding, an elementary module that draws much attention from researchers. A simple deep learning-based model is proposed by A. Karpathy \cite{karpathy2014large}, where a 2D CNN is independently applied to each video frame. To capture temporal information, the fusion of frame features is used, however, this simple approach cannot sufficiently capture the motion of objects. A straightforward approach to mitigate this problem is to directly introduce motion information into action recognition models. Hence, K. Simonyan \etal \cite{simonyan2014two} proposed a two-stream framework, where the spatial stream CNN takes a single frame as input and the temporal steam CNN takes a multi-frame optical flow as input. However, the model proposed by \cite{simonyan2014two} only employs shallow CNNs, while L. Wang \etal \cite{wang2016temporal} investigate different architectures and the prediction of a video is the fusion of the segments' predictions. To capture temporal information without introducing extra computational cost, J. Lin \etal \cite{lin2019tsm} propose a temporal shift module, which can fuse the information from the neighboring frames via using 2D CNNs. Another family of action recognition models is 3D based. D. Tran \etal \cite{tran2015learning} propose a deep 3D CNN, termed C3D, which employs 3D convolution kernels. However, it is difficult to optimize a 3D CNN, since there are much more parameters than in a 2D CNN. While J. Carreira \etal \cite{carreira2017quo} employs mature architecture design and better-initialized model weights, leading to better performance. All the above-mentioned models adopt the same frame rate, while C. Feichtenhofer \etal \cite{feichtenhofer2019slowfast} propose a SlowFast framework, where the slow path uses a low frame rate and the fast path employs a higher temporal resolution, and then the features are fused in the successive layers.

Recently, researchers pay more attention to video transformers -- a larger model with multi-head self-attention layers. G Bertasius \etal \cite{bertasius2021space} propose a video transformer model, termed TimeSformer, where video frames are divided into patches and then space-time attention is imposed on patches. TimeSformer has a larger capacity than CNNs, in particular using a large training dataset. Similarly, A. Arnab \etal \cite{arnab2021vivit} propose a transformer-based model, namely ViViT, which divides video frames into non-overlapping tubes instead of frame patches. Another advantage of the transformer-based model is that it is easy to use self-supervised learning approaches to pre-train the model. C. Wei \etal \cite{wei2021masked} propose a pre-training method, where the transformer-based model is to predict the \textit{Histograms of Oriented Gradients} (HoG) \cite{dalal2005histograms} of the masked patches and then the model is fine-tuned on downstream tasks. Using the pre-training-fine-tuning paradigm, we can further improve the performance of action recognition.

In this paper, we adopt multiple widespread action recognition and localization models to conduct experiments on our proposed dataset -- \TheName{}, showing that \TheName{} is relatively challenging for both action recognition and localization since the action is fast and the categories are imbalance (see sections \ref{sec:dataset} and \ref{sec:eval} for more details).

\section{Dataset} \label{sec:dataset}

\subsection{Dataset Construction}
\textbf{Preparation.} The procedure of data collection takes the following steps. First, we collect the raw broadcasting video clips of International/Chinese top-tier championships and tournaments from the ITTF (International Table Tennis Federation) Museum \& China Table Tennis Museum. We are authorized to download video clips from 2017 to 2021 (five years), including the Tokyo 2020 Summer Olympic table tennis competition, World Cups, World Championships, etc. 

As a convenience for further processing (e.g., storing and annotating), we split the whole game videos into 6-minute chunks, while ensuring the records are complete, distinctive, and of standard high-resolution, e.g., 720P ($1280\times 720$) and 1080p ($1920\times 1080$). Then, the short video chunks are ready for annotation in the next step.

\begin{figure}[ht]
\begin{center}
   \includegraphics[width=0.6\textwidth]{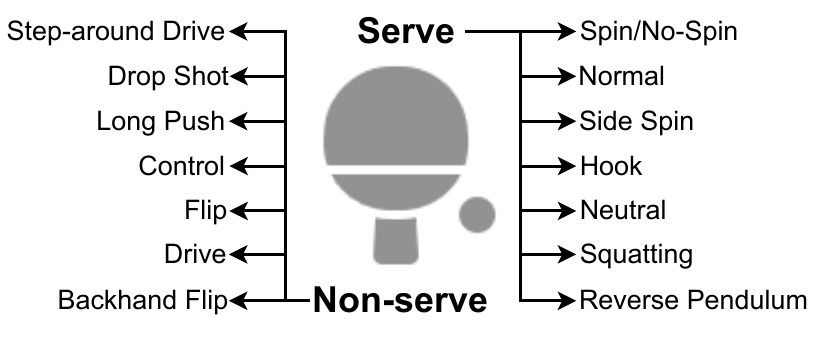}
\end{center}
   \caption{The classes of stroke/action in \TheName{} dataset.}
\label{fig:classes}
\end{figure}

\textbf{Annotation.} Since the actions (i.e., the ping pong strokes) in video chunks are extremely dense and fast-moving, it is challenging to recognize and localize all of them accurately. Thus, we cooperate with a team of table tennis professionals and referees to regroup the actions into broad types according to the similarity of their characteristics, which are the so-called 14-classes (14c) and 8-classes (8c). As the name implies, 14c categorizes all the strokes in our collected video chunks into 14 classes, which are shown in Fig.~\ref{fig:classes}. Compared to 14c, we further refine and combine the classes into eight higher-level ones, where the mapping relationships are revealed accordingly. Specifically, since there are seven kinds of service classes at the beginning of each game round (i.e., the first stroke by the serving player), we reasonably combine these service classes aside from those non-serving classes. Note that, all the actions/strokes we mentioned here are presented in the formal turns/rounds either conducted by the front-view player or back-view player from the broadcasting cameras (the blue-shirt and red-shirt player in Fig.~\ref{fig:camera}), and others presented in highlights (e.g., could be in a zoom-in scale with slow motions), Hawk-Eye replays (e.g., with the focus on the ball tracking), and game summaries (e.g., possibly in a zoom-out scale with a scoreboard) are ignored as redundant actions/strokes. For each of the action types in 14c/8c, we define a relatively clear indicator for the start frame and the end frame to localize the specific action/stroke. For example, we set the label of the non-serving stroke in a +/- 0.5 seconds window based on the time instance when the racket touches the ball, where the serving action/stroke may take a long time before the touching moment so that we label it in a +0.5/-1.5 seconds window as the adjustment. The window size is used as the reference for the annotator and can be varied according to the actual duration of the specific action. We further show the detailed distribution of the action/stroke classes in Section 3.2.

Once we have established the well-structured classes of actions, the annotators then engage to label all the collected broadcasting video clips. Unfortunately, the first edition of annotated dataset achieves around $85\%$ labeling precision under-sampling inspection by experts (i.e., international table tennis referees). It is understandable that, even with trained annotators, labeling the actions or strokes of table tennis players poses challenges due to 1) the fast-moving and dense actions; 2) the deceptive actions; 3) the entire/half-sheltered actions. To improve the dataset's quality, we collaborated with a team of table tennis professionals to correct inaccurate or missing samples from the dataset's first edition. The revised edition achieves an average labeling precision of approximately $95\%$, which is confirmed by the invited international table tennis referee~\cite{wufei}.

Note that, beyond labeling the segment of each action/stroke, we also label some additional information along with it (after finishing each action), which includes the placement of the ball on the table, the occasion of winning or losing a point, the player who committed the action (the name with the status of front view or back view in that stroke), whether it is forehand or backhand, and the serving condition (serve or not). The full list of labeling items is attached in the open-sourced dataset repository on Github\footnote{The link will be released later.}.

\textbf{Toolbox.}
To facilitate the above-mentioned annotating process, we design an efficient toolbox named BILS (\underline{B}aidu \underline{I}ntelligent \underline{L}abeling \underline{S}ystem) in \TheName{}. BILS is built upon a desktop-based application that enables the fast labeling process of videos and images, where the front-end and back-end modules leverage a large number of cutting-edge solutions for loading, interaction, temporal storage, and customization during the annotation. Compare to the existing open-sourced labeling tools, BILS is with the following advantages:
    \begin{figure}[ht]
\begin{center}
   \includegraphics[width=0.98\textwidth]{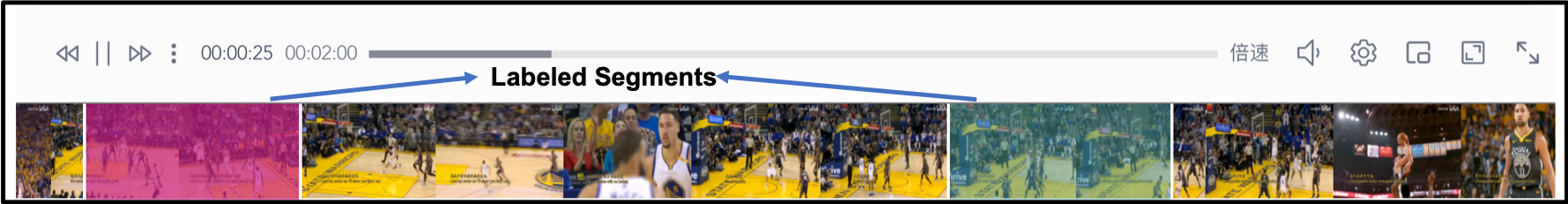}
\end{center}
   \caption{The timeline labeling in BILS.}
\label{fig:timeline}
\end{figure}
\begin{itemize}
    \item Timeline Labeling - the user can directly label actions/events on the thumbnail window by dragging the start/end point to locate the target moments or segments (e.g., the process is shown in Figure~\ref{fig:timeline}).

    \item Live Previewing - once the target moments or segments are labeled, BILS is able to provide a live preview of the expected output, while the user can adjust the time or reorganize the labels' format. Such fault-tolerable design can save a lot of time when the user confronts an infrequent mistake after finishing some repeated workload.
    \item Easy-to-Go Customizing - we adopt several efficiency-enhanced UI designs in BILS such as the customized label format (Figure~\ref{fig:label}), switchable themes, hot keys for labels, and multi-OS supports. BILS is not only limited to table tennis sports and is easy to adapt to any sport or even non-sport videos (e.g., basketball match videos in Fig.2\&3). 
\end{itemize}
Note that, the latest version of BILS with quick start and complete instruction documents are available in the same repository of \TheName{}, where a demo is also uploaded to showcase the full magic of BILS respectively. The innovation of the BILS can be seen as a significant contribution to the academic research field. BILS aims to enhance video and image labeling process in refining \TheName{}'s results. Calibration and live previewing are some of the features incorporated into BILS that improve the efficiency and precision of the annotation process.

    \begin{figure}[ht]
\begin{center}
   \includegraphics[width=0.75\textwidth]{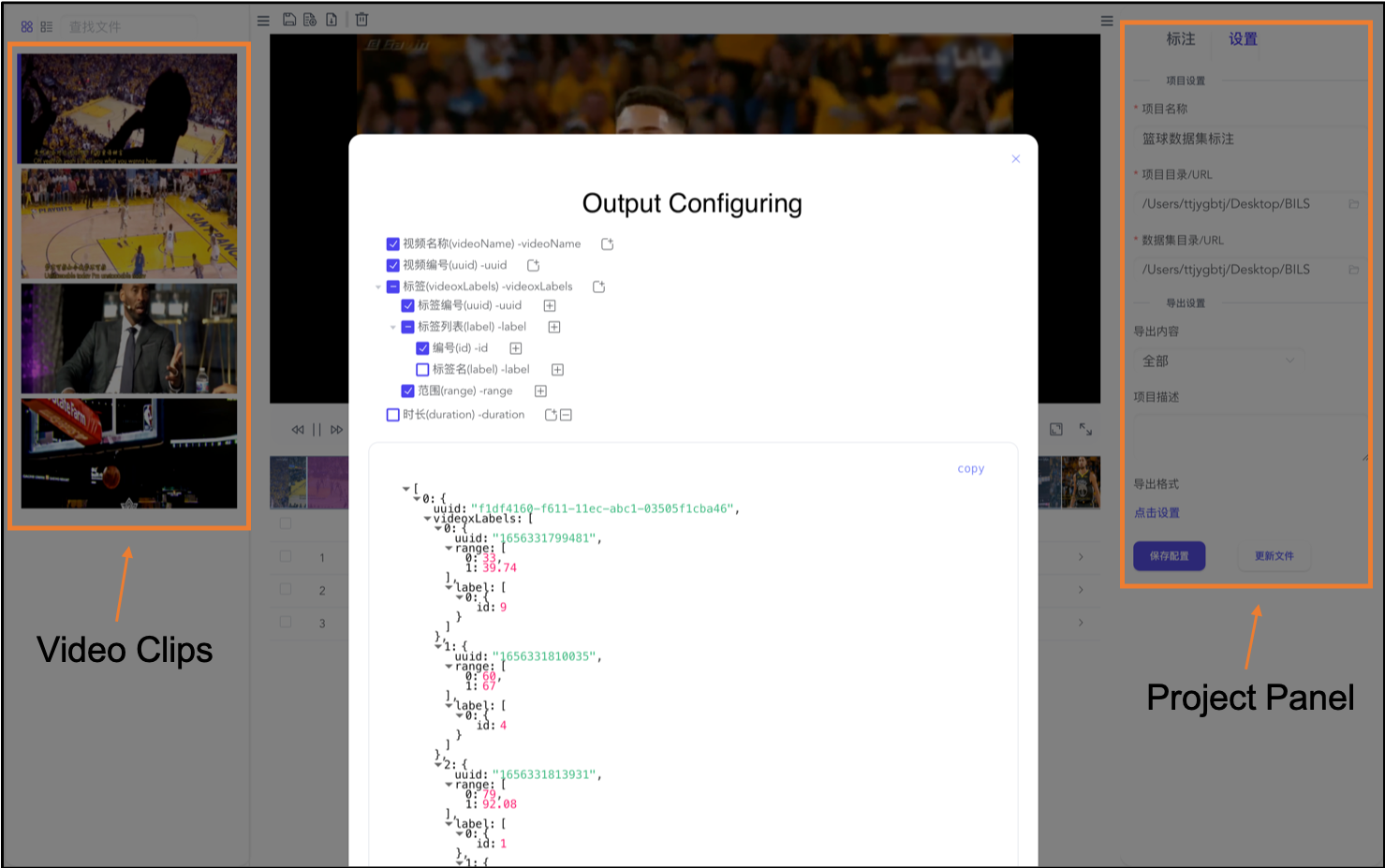}
\end{center}
   \caption{The customized label format in BILS.}
\label{fig:label}
\end{figure}

\renewcommand{\thempfootnote}{\fnsymbol{mpfootnote}}
\begin{table}[h]
\begin{minipage}{\textwidth}
\caption[xxx]{Comparison of existing sports video datasets for action detection and related purposes.}
\begin{center}
\scalebox{0.67}{
\begin{tabular}{|c|c|c|c|c|c|c|c|c|c|} 
 \hline
 Dataset & Duration & Samples & Segments & Classes & Tasks\footnote[2]{We denote the task of recognition, localization, and segmentation as ``Rec'', ``Loc'', and ``Seg''.} & Density\footnote[3]{We calculate the Density here as the value of Segments/Samples.} & Action Length & Source & Year \\ [0.5ex] 
 \hline\hline
 Olympic~\cite{niebles2010modeling} & 20 h & 800 & 800 & 16 & Rec & 1.00 & 5s $\sim$ 20s & YouTube & 2010\\ 
 \hline
 UCF-Sports~\cite{soomro2014action} & 15 h & 150 & 2,100 & 10 & Rec + Loc & 14.00 & 2s $\sim$ 12s & Broadcast TV & 2014\\ 
 \hline
 ActivityNet~\cite{caba2015activitynet}\footnote[1]{We also include some popular video-based action detection datasets, which are not only limited to sports-related actions but with a considerable amount of them in sub-genre.} & 648 h & 14,950 & 23,064 & 200 & Rec + Loc & 1.54 & 10s $\sim$ 100s & YouTube & 2015\\
 \hline
 TenniSet~\cite{faulkner2017tenniset} & 9 h & 380 & 3,568 & 10 & Rec + Loc & 9.40 & 1.08s $\sim$ 2.84s & YouTube & 2017 \\
 \hline
 TTStorke-21(MediaEval)~\cite{martin2018sport} & 94 h & 129 & 1,048 & 20 & Rec & 8.13 & 0.63s $\sim$ 2.27s & Self Recorded & 2018 \\
 \hline
 OpenTTGames~\cite{voeikov2020ttnet} & 5 h & 12 & 4,271 & 3 & Rec + Seg & 355.92 & 0.12s $\sim$ 0.13s & Self Recorded & 2020  \\
 \hline
 FineGym~\cite{shao2020finegym} & 708 h & 303 & 4,885 & 530 & Rec + Loc & 16.12 & 8s $\sim$ 55s & Broadcast TV & 2020 \\
 \hline \hline
 \textbf{\TheName{}} & 272 h & 2,721 & 139,075 & 14 &  Rec + Loc & 52.70 & 0.32s $\sim$ 3s & Broadcast TV & 2022 \\  
 \hline
\end{tabular}
\label{tab:comparison}}
\end{center}
\end{minipage}
\end{table}

\textbf{Calibration.} The last step for annotation is to clean the dataset for further research purposes. Since the whole 6-minute video dataset has some chunks without valid game rounds/turns (e.g., the opening of the game, the intermission and break, and the award ceremony), we filter out those unrelated chunks and focus on only action-related videos. Then, to accomplish the main goals of the research, which are tasks of \emph{recognition} and \emph{localization}, we further screen the video chunks to reserve the qualified ones meeting the following criteria:
\begin{itemize}[leftmargin=*]
    \item \textbf{\emph{Camera Angle}} -- The current most-common camera angle for broadcasting is the top view, where the cameras are hosted on a relatively high position upon the game court and overlook from the back of two sides of players. As shown in Fig.~\ref{fig:camera}, we only select the videos recorded in the standard top view and remove those with other views (e.g., side view, ``bird's eye'' view, etc.) to keep the whole dataset consistent. Note that, the broadcasting recordings with non-standard broadcasting angles rarely (less than $5\%$) appear in our target pool of competition videos, where only a few WTT (World Table Tennis Championships) broadcasting videos experimentally use the other types of camera angles. 
    \item \textbf{\emph{Game Type}} -- Considering the possible mutual interference in double or mixed double games, we only include the single games in \TheName{}. In this case, at most two players appear simultaneously in each frame, where usually one player is located near the top of the frame and another is at the bottom of the frame.
    \item \textbf{\emph{Audio Track}} -- Although we do not leverage the audio information for the localization and recognition tasks, the soundtracks are along with the videos and awaiting to be explored or incorporated in future research. We also remove those with broken or missing soundtracks.
    \item \textbf{\emph{Video Resolution}} -- We select the broadcasting videos with a resolution equal to or higher than 720P ($1280\times 720$) in MP4 format and drop those with low frame quality.
    \item \textbf{\emph{Action Integrity}} -- Since we trim the original broadcasting video into 6-minute chunks, some actions/strokes may be cut off when delivering the splitting. In this case, we make efforts to label those broken actions (about $0.2\%$) and it is optional to incorporate the broken actions or not in a specific task.
\end{itemize}

In the next subsection, we introduce the basic statistics of calibrated \TheName{} dataset and provide some guidance on pre-processing that might be helpful in specific tasks.  

\begin{figure}[ht]
\begin{center}
   \includegraphics[width=0.8\textwidth]{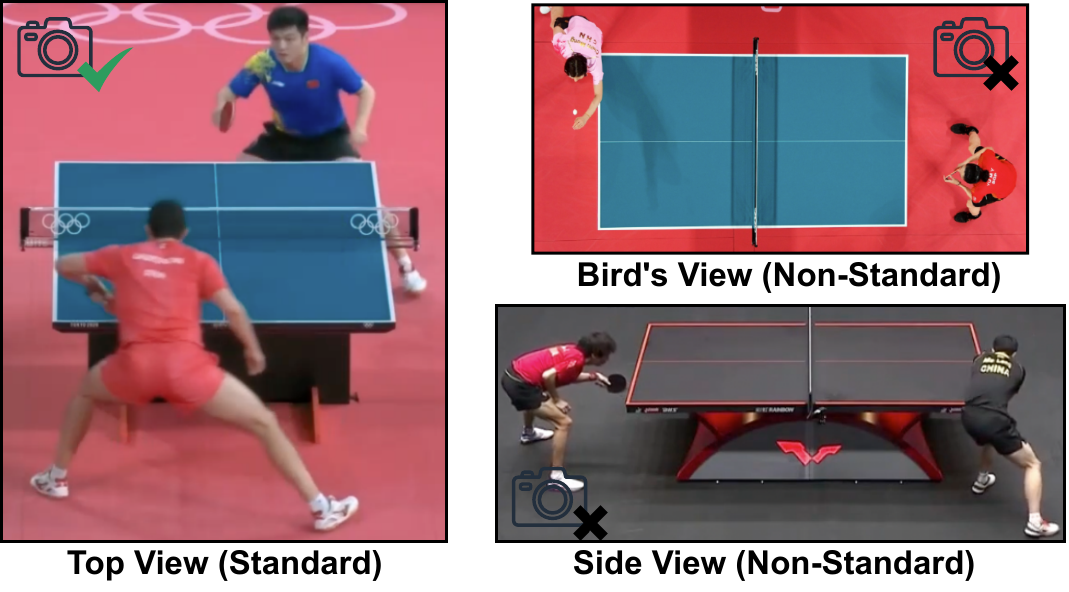}
\end{center}
   \caption{The camera angles in \TheName{} dataset.}
\label{fig:camera}
\end{figure}

\subsection{Dataset Analysis}
In this section, we conduct a comprehensive analysis of the statistics of \TheName{}, which is the foundation of the follow-up research and also serves as the motivation for establishing \TheName{} in the video analysis domain. The analysis lies in four parts, which are (1) general statistics, (2) category distribution, (3) densities, and (4) discussion. In the rest of this subsection, we go through each part in detail.

\noindent\textbf{General Statistics.} 
The \TheName{} dataset consists of 2,721 annotated 6-minute-long video clips, containing 139,075 labeled action segments, and last 272 hours in total. The video clips were extracted from over 200 table tennis competitions, involving almost all the top-tier ones during 2017-2021. These events include the World Table Tennis Championships, the Table Tennis World Cup, the Olympics, the ITTF World Tour, the Asia Table Tennis Championships, the National Games of China, as well as the China Table Tennis Super League. Since we intend to establish an action-focused table tennis dataset, the quality and standard of actions are expected to achieve a relatively high level, where the official HD broadcasting videos selected from the above-mentioned top-tier competitions lay a good foundation as the source of the target actions. 

\begin{figure*}[h!]
 \centering
 \includegraphics[scale=0.4]{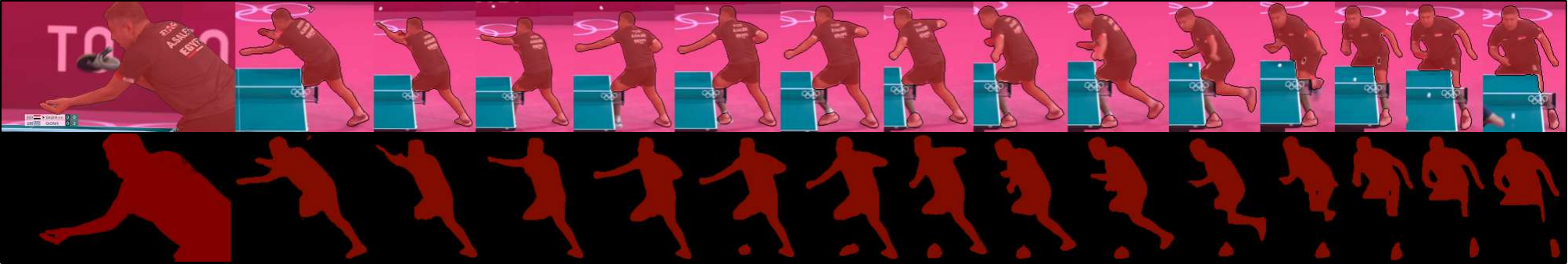}
 \caption{An example of ``Side Spin Serve'' action in the consecutive frames from \TheName{} dataset.}
 \label{fig:action_example}
\end{figure*}

\begin{figure*}[hpt!]
 \centering
 \includegraphics[scale=0.4]{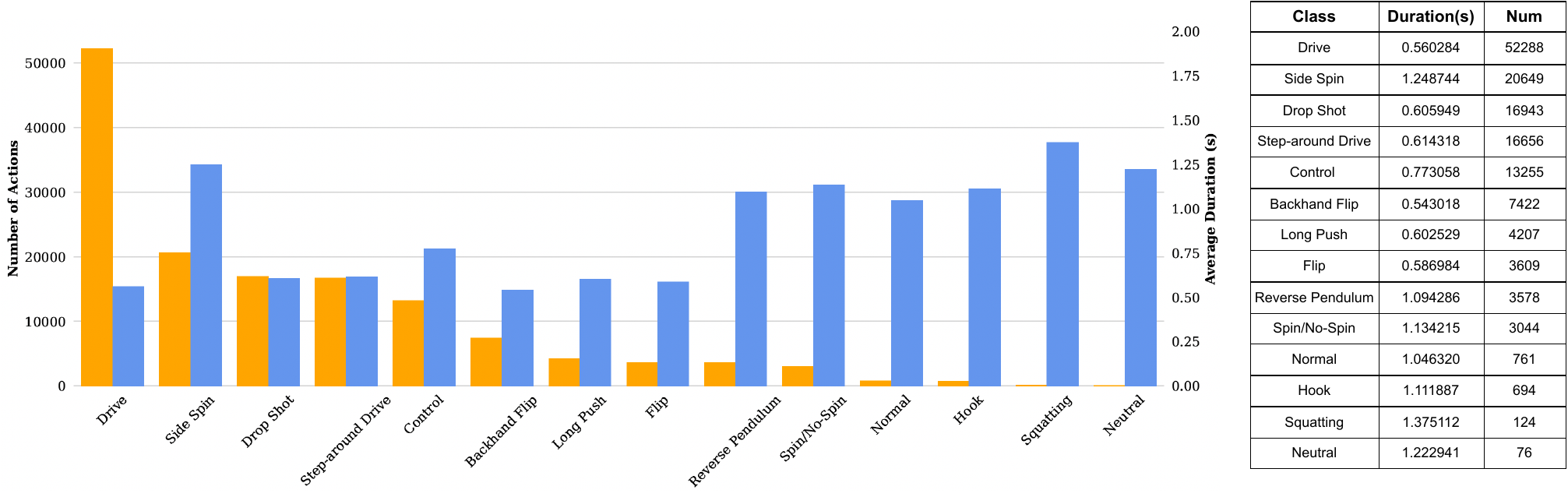}
 \caption{Sorted distribution of the number of actions (with duration) from each class (Ver. 14c) in the \TheName{} dataset.}
 \label{fig:num_cls}
\end{figure*}

\begin{figure*}[h]
\captionsetup[subfloat]{captionskip=2pt}
\centering
\subfloat[Serving Actions (7 classes)]{\includegraphics[width=0.33\textwidth]{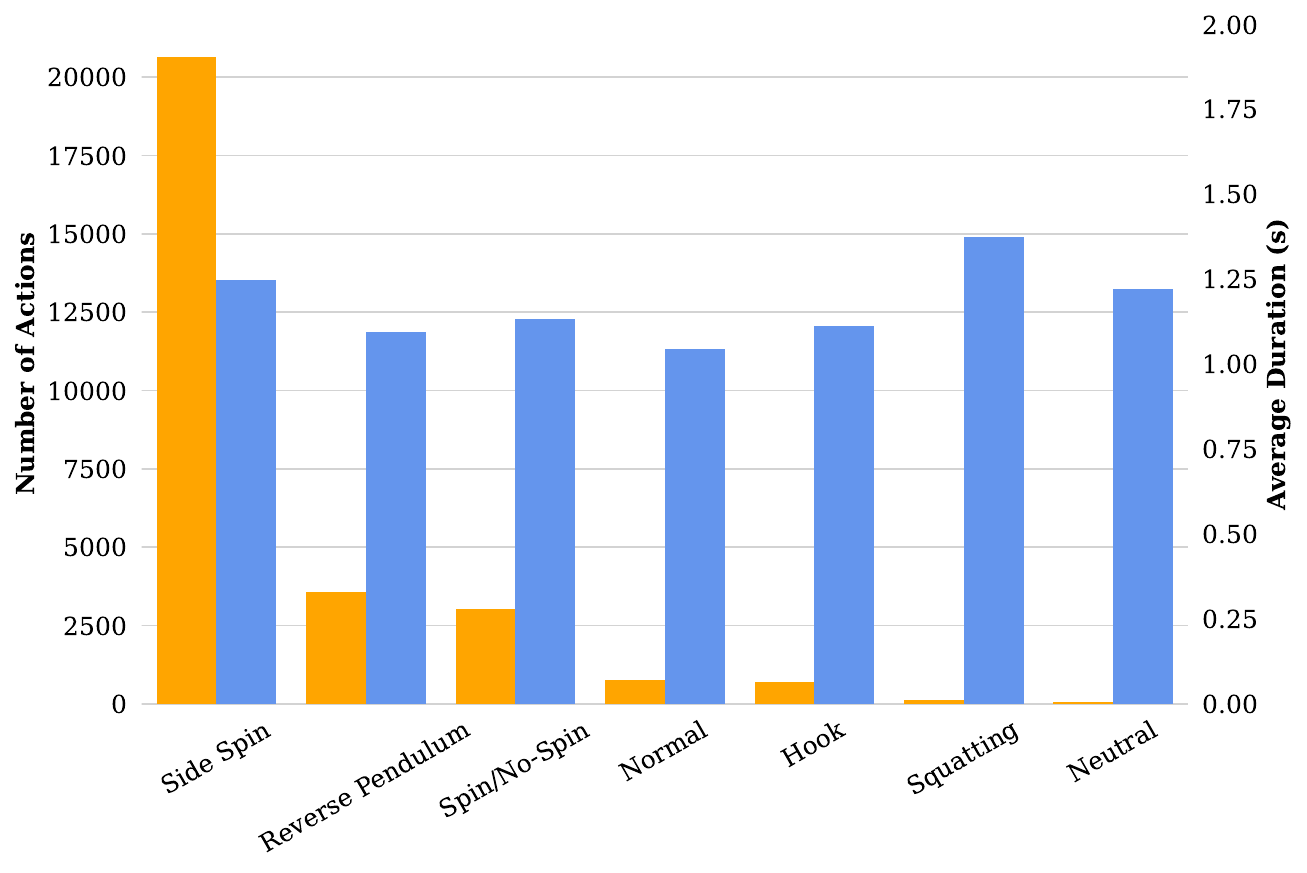}} 
\subfloat[Non-Serving Actions (7 classes) ]{\includegraphics[width=0.33\textwidth]{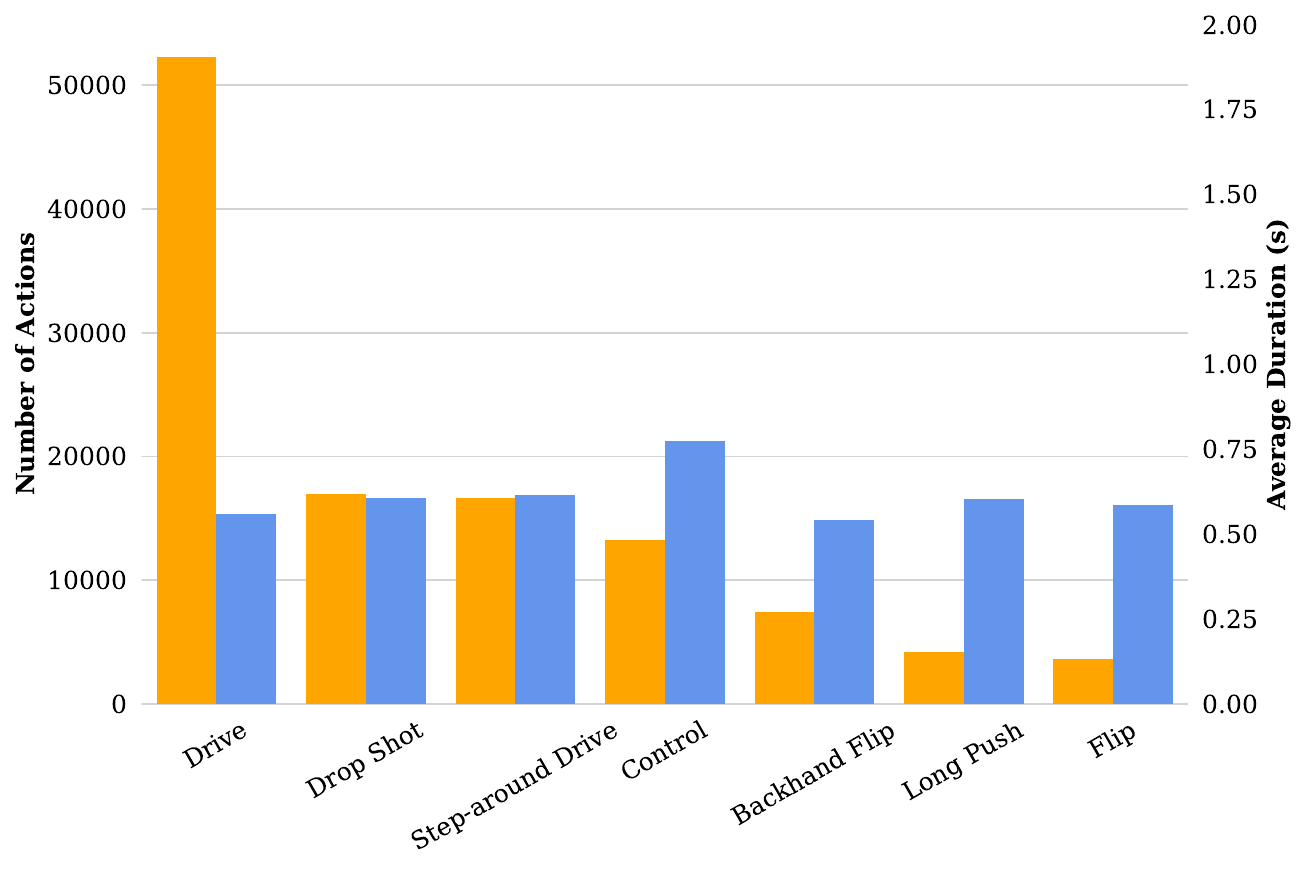}} 
\subfloat[Combined Actions (8 classes)]{\includegraphics[width=0.33\textwidth]{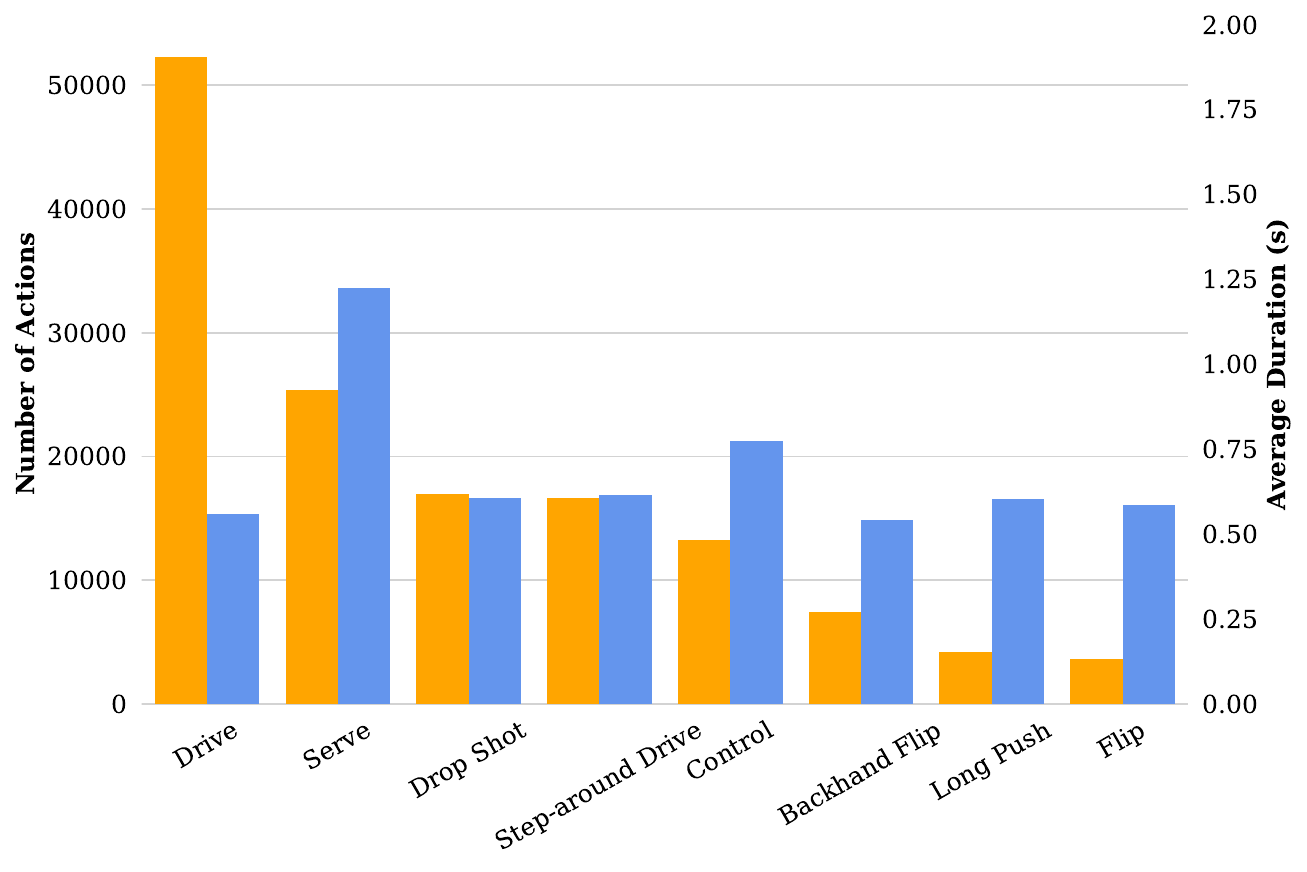}} 
\caption{Sorted distribution of the number of actions from the serving, non-serving, combined classes (Ver. 8c) in the \TheName{} dataset.}
\label{fig:two_version} 
\end{figure*}

For the general statistics of actions, Table~\ref{tab:comparison} presents the basic statistics of \TheName{} compared with the popular streams of sports-related datasets. The length of action for \TheName{} varies from 0.32 to 3 seconds, which is significantly shorter than most of the other datasets except the OpenTTGames and the TenniSet. Since the OpenTTGames targets only the in-game events (i.e. ball bounces, net hits, or empty event targets) without any stroking annotations, the action length is even shorter and with a considerably higher Density. Compared to these two table tennis-related datasets, our \TheName{} dataset has a longer Duration (over 30 times, and with more samples and segments), which means a lot for accurately reflecting the correlation and distribution of actions/strokes in different classes. Furthermore, compared to the sources of the OpenTTGames and the TenniSet, \TheName{} leverages the Broadcast TV games with the proper calibration by the international table tennis referee, which are more official and professional. Compared to the commonly used datasets such as ActivityNet and FineGym, \TheName{} focuses on the actions/strokes with dense and fast-moving characteristics. Specifically, the action length of \TheName{} is around 30 times shorter and density is 5 to 50 higher than these two datasets, which means \TheName{} introduces a series of action/strokes in totally different granularity. We later show that the SOTA models confront a huge challenge when applying on \TheName{}. Although the number of classes of \TheName{} is relatively small due to the nature of the table tennis sports itself, the action recognition and localization tasks are barely easy on \TheName{} for most of the mainstream baselines. In the next subsection, we show the distribution of categories in \TheName{}, which is one of the vital components causing the aforementioned difficulties.


\noindent\textbf{Category Distribution.}
Figure~\ref{fig:classes} shows the targeting categories of actions/stokes in \TheName{}. The \TheName{} dataset has two main branches of strokes -- \textbf{Serve} and \textbf{Non-serve}, where each of them owns 7 types of sub-strokes. Since the strokes are possibly taken by different players in the same video frames, we annotate strokes from both the front-view and the back-view players without time overlaps. For example, Figure~\ref{fig:action_example} represents consecutive frames of annotations on a front-view player, where a player is producing a ``Side Spin Serve'' at the start of a turn. Notice that this player stands at the far-end of the game table and turns around from the side-view to the front-view, then keeps the front-view for the rest of this turn. A similar annotating is taken to record the player's strokes at the near-end of the game table.

As aforementioned, the \TheName{} dataset consists of two versions, which are 14c (14 classes) and 8c (8 classes). The 14c flattens all the predefined classes in Figure~\ref{fig:camera} and treats them equally in future tasks. We measure the number of strokes for each of the classes in Figure~\ref{fig:num_cls}. Overall, the drive category contains more than half of the samples (we sort the categories in descending order), which means there exists unbalancing category phenomenon in \TheName{}. We also observe an interesting point that the non-serve categories generally last a shorter duration than the serve categories (as shown from the blue bars). This is because of the long preparation of the serve strokes and could be an important feature to distinguish the serve and the non-serve strokes. Another fact is that the non-serve categories dominate the whole dataset, where there are seven non-serve categories out of the left-most eight categories (in descending order). This unbalancing phenomenon leads to the creation of the second version -- 8C to combine all the serve categories into one unified ``Serve'' category.

Figure~\ref{fig:two_version} measures the number of actions in the serve, non-serve, and combined 8c categories separately. As a result, the sub-categories are unbalanced, where ``Side Spin'' and ``Drive'' dominates the serving and non-serving actions in (a)\&(b). In the combined 8c dataset, ``Drive'' unsurprisingly takes a large proportion among all eight categories due to its frequent appearances in modern table tennis competitions. Thus, it is necessary to adopt additional sampling techniques to mitigate the side effects, and the implementation details are introduced in Section 4.

\begin{figure}[h]
\captionsetup[subfloat]{captionskip=2pt}
\centering
\subfloat[All Classes]{\includegraphics[width=0.51\textwidth]{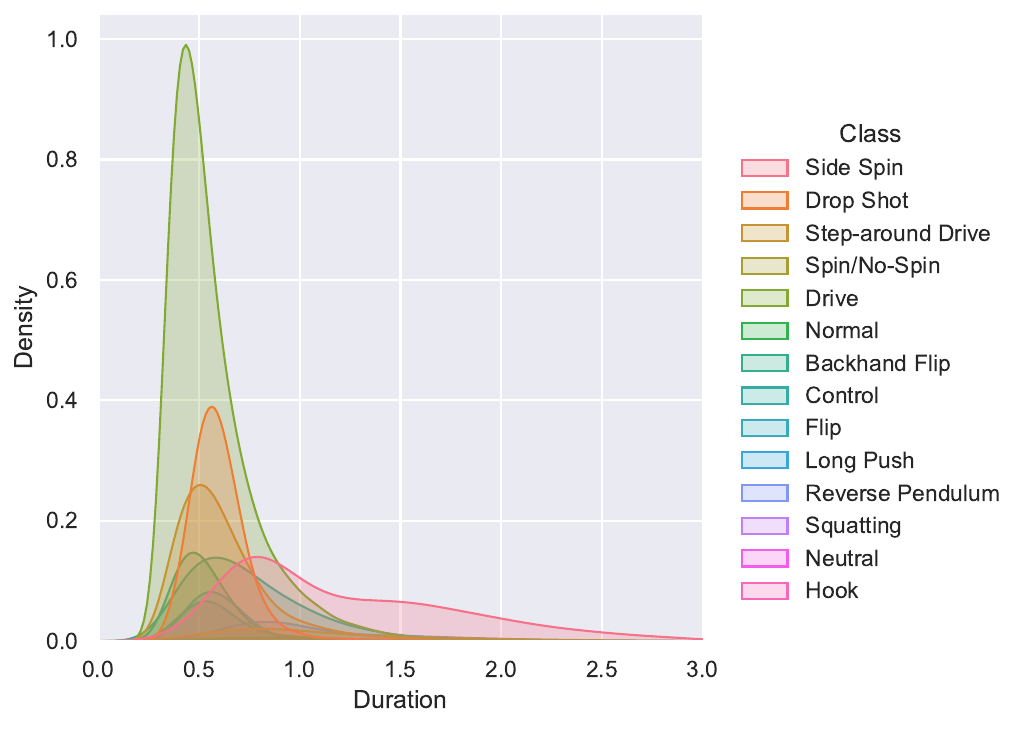}} 
\subfloat[Serve vs Non-serve Classes]{\includegraphics[width=0.47\textwidth]{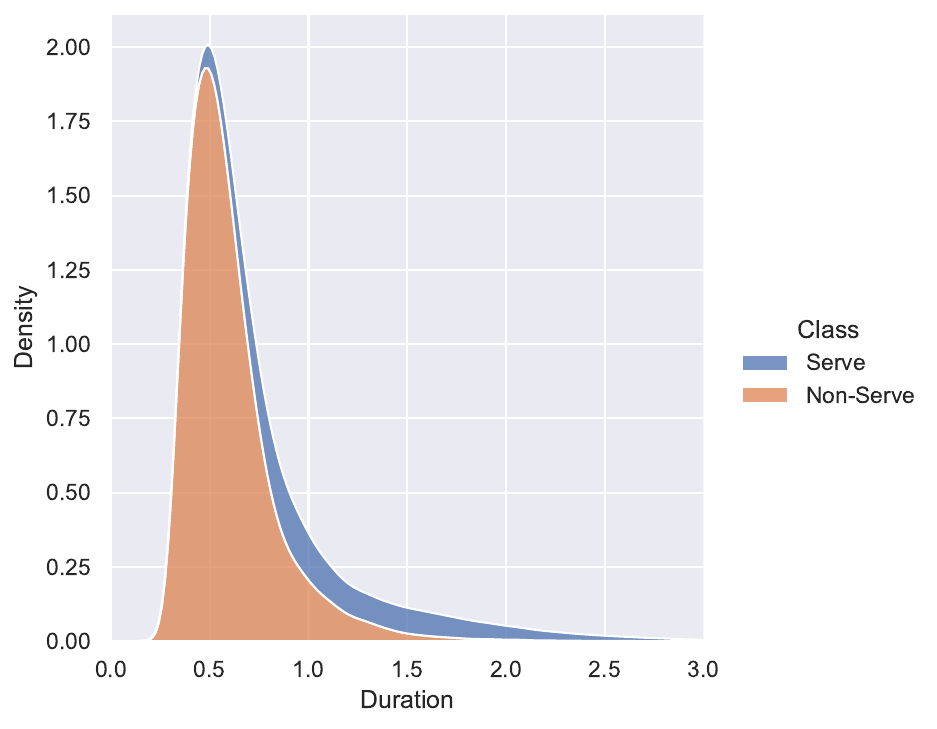}} 
\caption{Time Duration distribution of actions/strokes in \TheName{} dataset.} 
\label{fig:dist} 
\end{figure}

\noindent\textbf{Densities.}
As one of the unique characteristics compared to other datasets, the high density of actions/strokes plays a vital role in \TheName{}. We analyze the density of actions/strokes in two aspects, where the first one is the duration distribution of each action/stroke category. Figure~\ref{fig:dist}(a) shows the duration distribution of all classes of strokes in \TheName{}. We can observe that most of the action types have relatively stable distributions (with short tails) except the ``Control'' and the ``Neutral'' types. The ``Neutral'' is one of the serving strokes representing those irregular/unknown actions in the serving process, where the duration of it widely ranges from 0.4 to 3 seconds in Figure~\ref{fig:dist}(a). It is similar that the ``Control'' category stands for the non-serving strokes which cannot be categorized into any other classes, and has a long tail of 2 seconds at most. On the whole, the average duration of strokes sticks to around 0.5 seconds, which demonstrates the actions in \TheName{} is fast-moving compared to other datasets in Table 1. We also compare the duration distribution between the serve and non-serve classes in Figure~\ref{fig:dist}(b), where the stacked plot of both classes presents that the serve class has a slightly longer tail than the non-serve class.

\begin{figure}[h]
\captionsetup[subfloat]{captionskip=2pt}
\centering
\subfloat[The Action Frequency]{\includegraphics[width=0.453\textwidth]{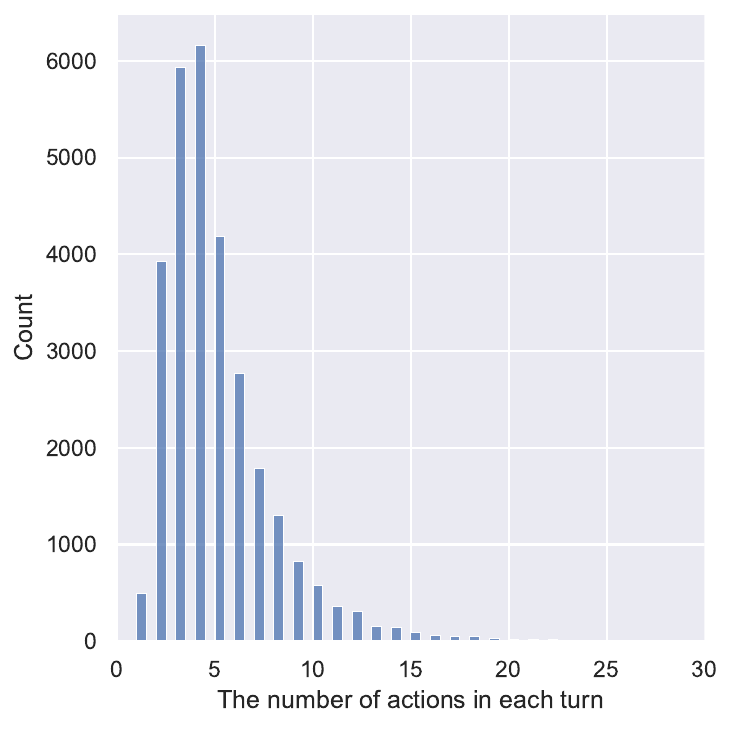}} 
\subfloat[The Appearance Density ]{\includegraphics[width=0.46\textwidth]{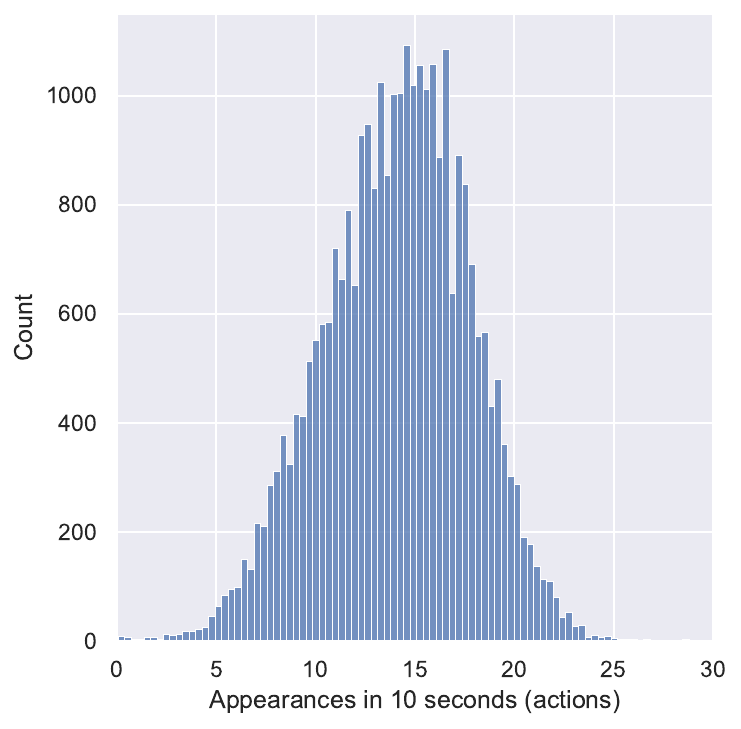}} 
\caption{Density of actions/strokes in each turn of the game.} 
\label{fig:dense} 
\end{figure}

For the second respect, we measure the action frequency and the appearance density at each turn of the game. One turn in a game starts from a serving action by one player and ends with one point acquired by either of the players. Thus, the action frequency can reflect the nature of the table tennis competition (e.g., the strategy and the pace of the current table tennis game) and how many strokes could be delivered in a single turn. As shown in Figure~\ref{fig:dense}(a), most of the turns have around 3 $\sim$ 5 actions/strokes and the distribution is also long-tailed to a maximum of 28 actions/strokes per turn. It is reasonable that the pace of the game today becomes faster than ever, and a point could be outcome within three strokes for both players (i.e., summed up within 6 strokes). To further differentiate from other sports datasets, we additionally measure the appearance density in 10 seconds, which is usually the time boundary for long-duration and short-duration sports actions. Figure~\ref{fig:dense}(b) depicts the histogram of the action counts in 10-second consecutive video frames. The results follow a normal-like distribution with average counts as 15, which means the actions appear densely and frequently in a short time in the most of the time in \TheName{}. The density can be achieved as about 1.5 actions per second on average and it demonstrates the characteristics of dense actions for the \TheName{} dataset.

\noindent\textbf{Summary.} 
In this section, we comprehensively report the statistics of \TheName{}. Specifically, we leverage rich measurements to have a thorough view of the volume of the dataset, the category distribution, the duration of samples, and the densities. In conclusion, the \TheName{} dataset is unique compared to the other mainstream sports datasets in the following sides,
\begin{itemize}[leftmargin=*]
    \item Comparing to the table tennis datasets~\cite{voeikov2020ttnet, faulkner2017tenniset}, \TheName{} has an adequate sample size (10 times $\sim$ 200 times) and more subtle set of categories.  
    \item Comparing to the large datasets in other sports domain~\cite{niebles2010modeling,soomro2014action,caba2015activitynet,shao2020finegym}, \TheName{} focuses on the dense (5 times $\sim$ 50 times) and fast-moving (around 0.1 times) actions.
    \item Comparing to most of the aforementioned sports datasets, \TheName{} includes two players' actions back-to-back (the action appears in turns with a fast pace), which is more complicated and mutually interfering when analyzing. 
\end{itemize}


\section{Benchmark Evaluation}\label{sec:eval}
In this section, we present the evaluation of various methods on \TheName{} dataset and show the difficulties when applying the original design of those methods. 

\begin{table}[htp!]
    \centering
    \caption{Key parameters of implemented algorithms.}
    \scalebox{0.7}{
    \begin{tabular}{c|c|c|c|c|c}
    \toprule
        Recognition Method & Backbone & Sampling & Num\_Seg & Target Size & Others \\
        \midrule
        TSN & ResNet50-D & TenCrop & 8 & 224 & Distillation with ResNet152-CSN\\ 
        TSM & ResNet50-D & Dense & 8 & 224 & Distillation with ResNet152-CSN\\
        SlowFast & ResNet50-D & Time-stride & 8 & 224 & Mutigrid speedup~\cite{wu2020multigrid}\\
        TimeSformer & ViT-base-patch16-224 & UniformCrop & 8 & 224 & Linspace when testing\\
        ViD & SwinT-base-1k-224 & UniformCrop & 32 & 224 & -\\
        Attention-LSTM & Inception-ResNet-v2 & RandomNoise & 8 & 224 & -\\
        MoViNet & MoViNet-A0 & UniformCrop & 50 & 172 & AutoAugment~\cite{cubuk2018autoaugment}\\
        VideoMAE V2-g & ViT-giant & MultiScaleCrop & 8 & 224 & Post-pre-training\\
         \midrule
         Localization Method & Backbone & Sampling & NMS & Window Size & Others \\
         \midrule
         BSN & TSM (ResNet50-D) & - & Soft &100  & - \\
         BSN++ & TSM (ResNet50-D) & - & Soft &100 & Proposal extension~\cite{su2020bsn++}, Self-attention \\
         SSTAP & TSM (ResNet50-D) & - & Soft &100 & Self-Supervised Learning\\
         BMN  & TSM (ResNet50-D) & - & Soft &100& Proposal extension~\cite{su2020bsn++} \\
         TCANet+  & TSM (ResNet50-D) & RandomNoise & Soft & 300 & LGTE, SEBlock~\cite{qing2021temporal}\\
         \bottomrule
    \end{tabular}
    \label{tab:parameters}}
\end{table}

\subsection{Baselines}
We adopt two groups of baseline algorithms to separately tackle the action recognition and action localization tasks. For \textbf{Action Recognition}, we include the trending algorithms as follows.
\begin{itemize}[leftmargin=*]
    \item Temporal Segment Network (TSN)~\cite{wang2016temporal} is a classic 2D-CNN-based solution in the field of video action recognition. This method mainly solves the problem of long-term behavior recognition of video and replaces dense sampling with sparsely sampling video frames, which can not only capture the global information of the video but also remove redundancy and reduce the amount of calculation. 

    \item Temporal Shift Module (TSM)~\cite{lin2019tsm} is a popular video action recognition model with shift operation, which can achieve the performance of 3D CNN but maintain 2D CNN’s complexity. The method of moving through channels greatly improves the utilization ability of temporal information without increasing any additional parameters and calculation costs. TSM is accurate and efficient: it ever ranked first place on the Something-Something~\cite{goyal2017something} leaderboard. Here, we adopt the industrial-level variation from the deep learning platform PaddlePaddle~\cite{ma2019paddlepaddle}. We improve the original design with additional knowledge distillation, where the model is pretrained on Kinetics400~\cite{carreira2017quo} dataset.
    
    \item SlowFast~\cite{feichtenhofer2019slowfast} involves a Slow pathway operating at a low frame rate to capture spatial semantics, and another Fast pathway operating at a high frame rate to capture motions in the temporal domain. SlowFast is a powerful action recognition model, which ranks in second place on AVA v2.1 datasets~\cite{gu2018ava}.
    
    \item Video-Swin-Transformer (Vid)~\cite{liu_2021} is a video classification model based on Swin Transformer~\cite{liu2021swin}. It utilizes Swin Transformer's multi-scale modeling and the efficient local attention module. Vid shows competitive performances in video action recognition on various datasets, including the Something-Something and Kinetic-series datasets.
    
    \item Attention-LSTM~\cite{long2018attention} uses a two-way long and short-term memory network with attention layers to encode all the frame features of the video in sequence.
    
    \item MoViNet~\cite{kondratyuk2021movinets} is a mobile video network developed by Google research. It uses causal convolution operators with stream buffer and temporal ensembles to improve the accuracy of video classification. It is a lightweight and efficient video model designed for online video stream reasoning.
    
    \item TimeSformer~\cite{bertasius2021space} is a video classification model based on a vision transformer, which equips with the global receptive field, and strong time series modeling without convolutions. At present, it has achieved SOTA accuracy on the Kinetics-400 data set, approximating the performance of classic 3D-CNN-based video classification models, while it has a considerably short training time.

    \item VideoMAE V2-g~\cite{wang2023videomae} emerges as a groundbreaking advancement that redefines the boundaries of video foundation models. Building upon the foundational principles of Masked Autoencoders with powerful ViT-giant backbone, this cutting-edge model introduces a revolutionary concept: Dual Masking. With this innovative approach, VideoMAE V2 achieves unprecedented scalability, efficiency, and creativity in building video foundation models for recognition.
\end{itemize}

For \textbf{Action Localization}, we investigate the following classical or trending algorithms,
\begin{itemize}[leftmargin=*]
    \item Boundary Sensitive Network (BSN)~\cite{lin2018bsn} is an effective proposal generation method, which adopts a "local to global" fashion. BSN has already achieved high recalls and temporal precision on several challenging datasets, such as ActivityNet-1.3~\cite{caba2015activitynet} and THUMOS14'~\cite{wang2014action}.
    \item BSN++~\cite{su2020bsn++} is a new framework that exploits complementary boundary regressor and relation modeling for temporal proposal generation. It ranked first in the CVPR19 - ActivityNet challenge leaderboard on the temporal action localization task.
    \item Boundary-Matching Network (BMN)~\cite{lin2019bmn} introduces the Boundary-Matching (BM) mechanism to evaluate confidence scores of densely distributed proposals, which leads to generating proposals with precise temporal boundaries as well as reliable confidence scores simultaneously. Combining with existing feature extractors (e.g., TSM), BMN can achieve state-of-the-art temporal action detection performance.
    \item Self-Supervised Learning for Semi-Supervised Temporal Action Proposal (SSTAP)~\cite{wang2021self} is one of the self-supervised methods to improve semi-supervised action proposal generation. SSTAP leverages two crucial branches, i.e., the temporal-aware semi-supervised branch and the relation-aware self-supervised branch to further refine the proposal-generating model.
    \item Temporal Context Aggregation Network (TCANet)~\cite{qing2021temporal} is the championship model in the CVPR 2020 - HACS challenge leaderboard, which can generate high-quality action proposals through "local and global" temporal context aggregation and complementary as well as progressive boundary refinement. Since it is a newly released model designed for several specific datasets (the original design shows unstable performance on \TheName{}), we slightly modify the architecture to incorporate a BMN block ahead of TCANet. The revised model is denoted as TCANet+ in our experiments.
\end{itemize}

\subsection{Experimental Setup}

\textbf{Datasets.} As aforementioned, we establish two versions -- 14-classes (14c) and 8-classes (8c) -- in \TheName{}. The difference is that we combine all the serving strokes/actions in a single ``Serve'' category in the 8c version. We evaluate the baseline action recognition algorithms on both versions and compare the performances side by side. Since the performances of the current action localization algorithms on 14c are far from satisfactory, we only report the results on 8c for reference.

\textbf{Metrics.} We use the Top-1 and Top-5 accuracy as the metric for evaluating the performance of baselines in the action recognition tasks. For the action localization task, We adopt the area under the Average Recall vs. Average Number of Proposals per Video (AR-AN) curve as the evaluation metric. A proposal is a true positive if it has a temporal intersection over union (tIoU) with a ground-truth segment that is greater than that or equal to a given threshold (e.g, tIoU $> 0.5$). AR is defined as the mean of all recall values using tIoU between 0.5 and 0.9 (inclusive) with a step size of 0.05. AN is defined as the total number of proposals divided by the number of videos in the testing subset. We consider 100 bins for AN, centered at values between 1 and 100 (inclusive) with a step size of 1 when computing the values on the AR-AN curve.

\begin{figure}[h]
\captionsetup[subfloat]{captionskip=2pt}
\centering
\subfloat[Results \emph{\underline{without}} class balancing]{\includegraphics[width=0.49\textwidth]{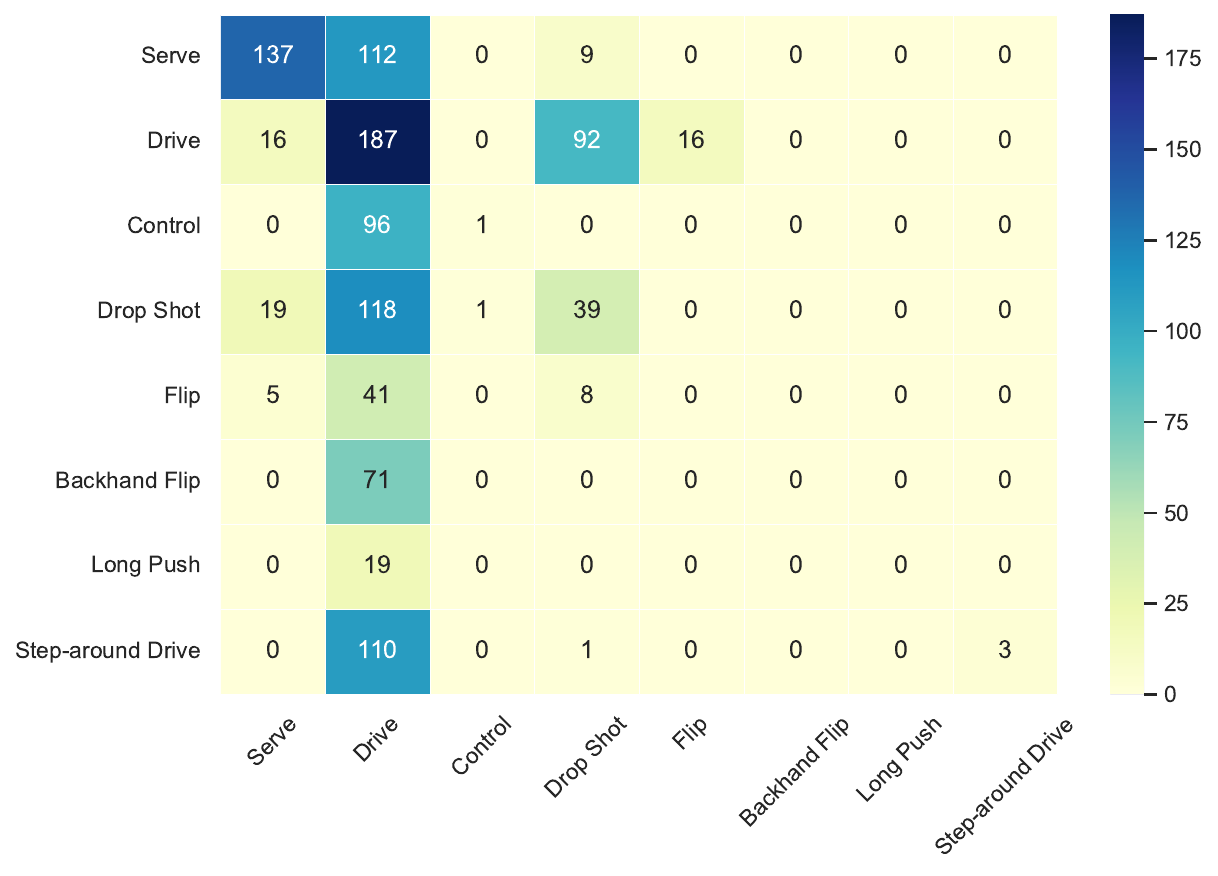}}  \ 
\subfloat[Results \emph{\underline{with}} class balancing]{\includegraphics[width=0.49\textwidth]{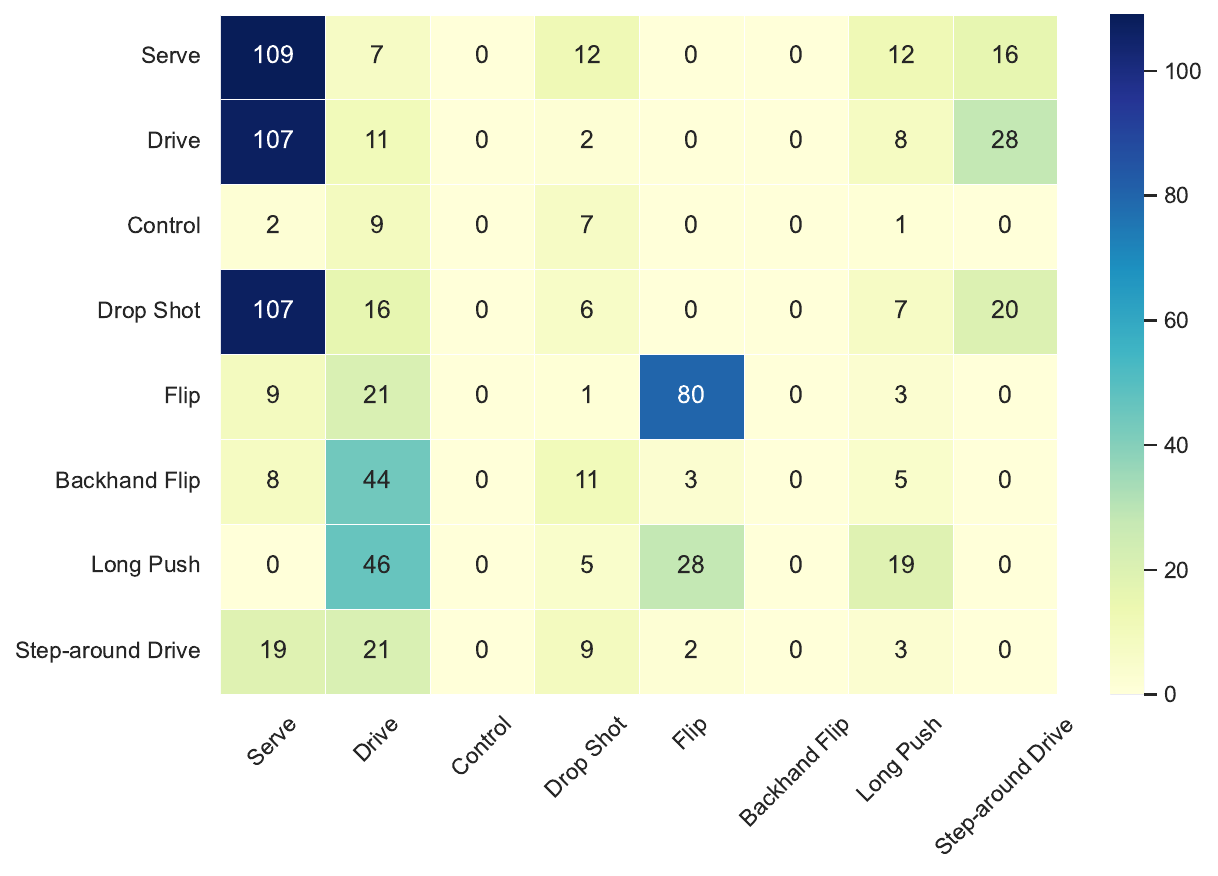}} 
\caption{Confusion matrix of action recognition results} 
\label{fig:heat} 
\end{figure} 

\begin{figure}[!t]
 \centering
 \includegraphics[scale=0.33]{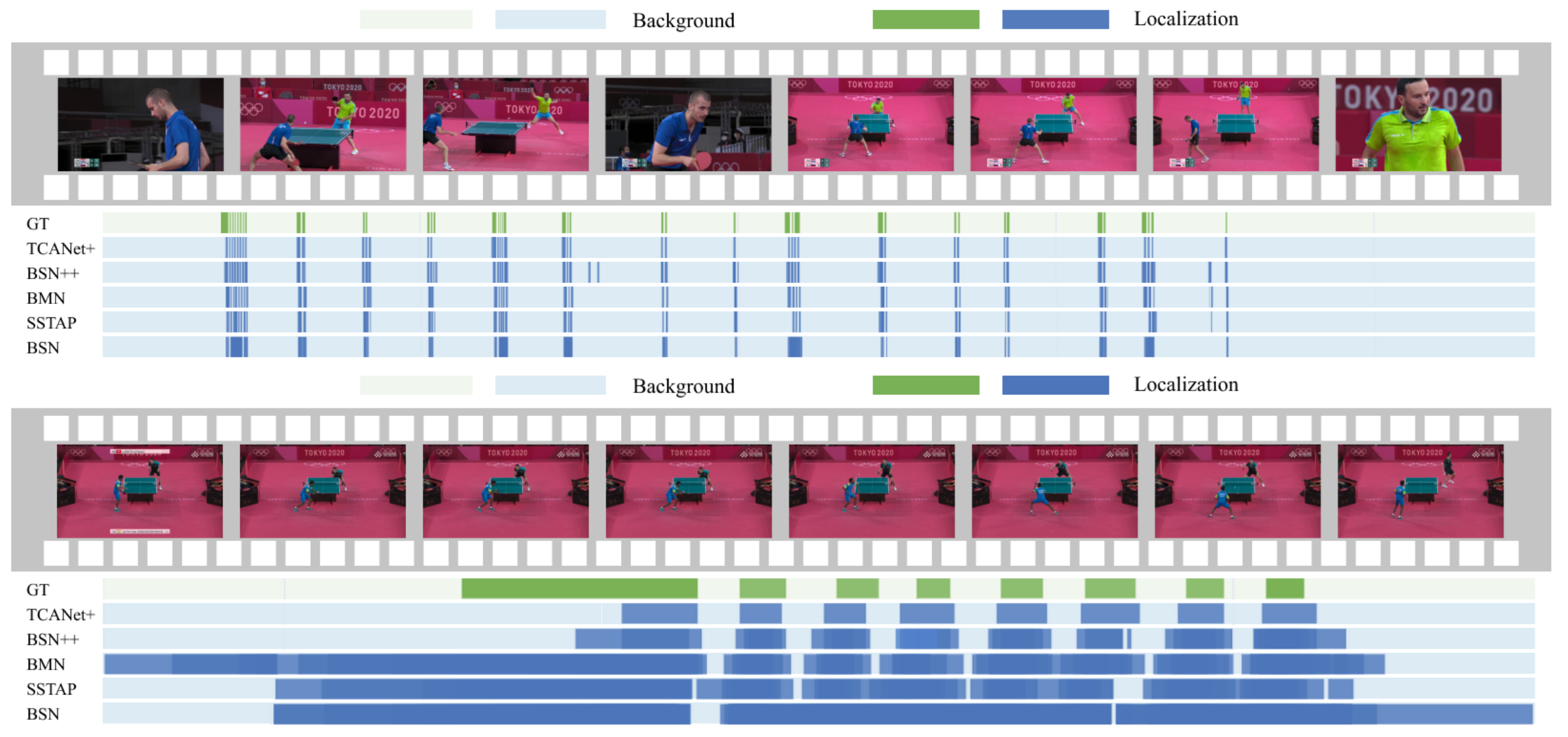}
 \caption{Example of localisation results on \TheName{} dataset.}
 \label{fig:predict}
\end{figure}

\textbf{Data Augmentation.} 
The category distribution analysis in Section 3.2 reveals that \TheName{} is an imbalanced dataset, which might cause trouble to the recognition task. To fully utilize the proposed \TheName{} dataset, we design a simple yet effective data augmentation method to train the baseline models. Specifically, we introduce an up/down-sampling procedure to balance the sample size of each category. First, we calculate the mean size of all the samples (denoted $N$ samples) by the defined number of categories (e.g., 8c or 14c) as $M$, where $M = N/8 $ or $N/14$. Then, we sample the actual number of action segments $S_A$ within a range $M \leq S_A \leq 2M$. For those categories with less number of segments than $M$, we up-sample by the random duplication. For those with much more samples over $2M$, we random down-sample from the original sample set and try to make the sampled actions cover all the video clips (i.e., uniform sampling from each 6-minute chunk). We apply this data augmentation strategy to all the recognition tasks and the significant performance gains. As shown in Figure~\ref{fig:heat}, we report the ablation results of action recognition baseline TSM on \TheName{} with/without the data augmentation. In Figure~\ref{fig:heat}(a), the confusion matrix illustrates that TSM falsely classifies most of the strokes/actions into the ``Drive'' category without the class balancing, while this kind of misclassification is significantly alleviated after we apply the data augmentation. Note that, since the localization tasks normally require feature extractor networks as the backbones (e.g., TSM is one of the backbones), localization performance could also be benefited from the designed data augmentation. The baseline algorithms in the following results section also involve data augmentation.




\textbf{Implementations.}
We list the key hyper-parameters of all the implemented action recognition and action localization algorithms in Table~\ref{tab:parameters}. Note that, the TSN and TSM stand for the revised versions compared to the released version from the original paper. For other baselines, we follow the original design of the critical components from each algorithm and adjust specific hyper-parameters to fit them on the \TheName{} dataset.

\subsection{Results}


\begin{figure}[h]
\captionsetup[subfloat]{captionskip=2pt}
\centering
\subfloat[BSN]{\includegraphics[width=0.32\textwidth]{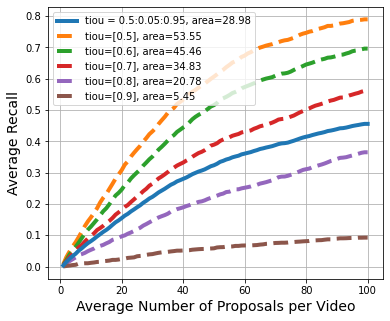}}\ 
\subfloat[BSN++]{\includegraphics[width=0.32\textwidth]{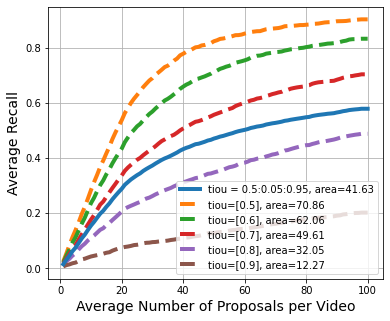}}\ 
\subfloat[SSTAP]{\includegraphics[width=0.32\textwidth]{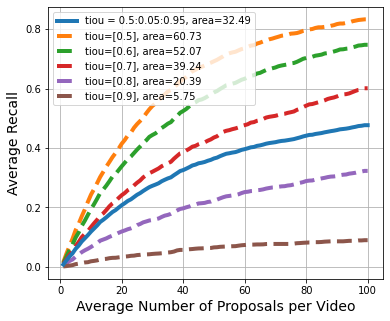}}\ \\
\subfloat[TCANet+]{\includegraphics[width=0.32\textwidth]{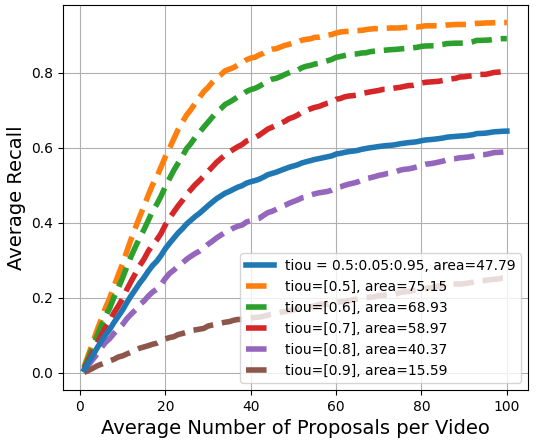}}\ 
\subfloat[BMN]{\includegraphics[width=0.32\textwidth]{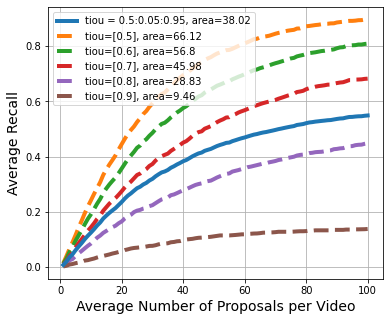}}\
\subfloat[ALL]{\includegraphics[width=0.32\textwidth]{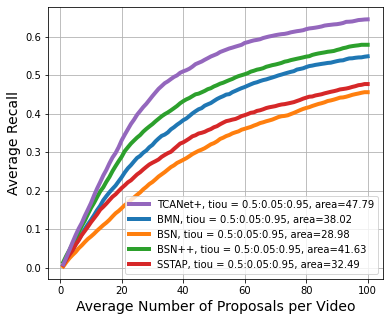}}
\caption{Performance of different action localization methods on \TheName{} (Best Viewed with 200\% Zoom-in).} 
\label{fig:fig1} 
\end{figure} 

\begin{table}[ht!]
\caption{Performance of different action recognition methods on \TheName{} in terms of validation accuracy.}
    \centering
    \scalebox{0.82}{
    \begin{tabular}{c|c|cc|cc}
        \toprule
        Actions & Method & Acc1@14c & Acc5@14c & Acc1@8c & Acc5@8c \\ 
         \hline
        \multirow{9}*{Face} & 
         TSN & 59.77 & 90.13 & 73.84 & 94.85\\
        & Vanilla-TSN & 57.28 & 89.36 & 72.07 & 93.89 \\ 
        & TSM & 69.97 & 98.08 & 82.35 & \textbf{99.06}\\
        & Vanilla-TSM & 63.33 & 93.21 & 76.79 & 95.63 \\
        & SlowFast & 63.19 & 92.79 & 77.41 & 96.07\\
        & Vid & 68.85 & \textbf{98.16} & 81.03 & 98.96\\
        & Attention-LSTM & 53.19 & 81.98 & 73.34 & 89.31 \\ 
        & MoViNet & 54.79 & 85.31 & 74.09 & 88.93 \\
        & TimeSformer & 64.01 & 94.87 & 77.39 & 95.56 \\
        & VideoMAE V2-g & \textbf{74.23} & 98.15 & \textbf{85.03} & 98.97 \\
        \midrule
        \multirow{9}*{Face\&Back} & 
         TSN & 53.31 & 78.03 & 54.13 & 82.46 \\
        & Vanilla-TSN & 50.47 & 76.69 & 53.94 & 80.31\\ 
        & TSM & 59.46 & 87.11 & 68.39 & 91.73 \\
        & Vanilla-TSM & 55.72 & 84.05 & 63.39 & 86.32 \\
        & SlowFast & 56.39 & \textbf{87.98} & 65.10 & 89.93 \\
        & Vid & 58.54 & 87.20 & 67.28 & 91.05\\
        & Attention-LSTM & 51.06 & 75.39 & 52.79 & 80.67 \\ 
        & MoViNet & 52.45 & 77.19 & 54.12 & 82.07 \\
        & TimeSformer & 56.49 & 86.31 & 64.93 & 90.30 \\
        & VideoMAE V2-g & \textbf{61.21} & 87.05 & \textbf{73.83} & \textbf{93.41} \\
        \bottomrule
    \end{tabular}}
    \label{tab:recognition}
\end{table} 

This section first presents the performance results of action recognition tasks. Table~\ref{tab:recognition} provides a summary of these results in terms of top1/top5 accuracy on both 8c and 14c datasets. As we can observe, the top5 accuracy is unsurprisingly higher than the top1 accuracy for all the methods. For comparing the 8c and 14c datasets in the same setting, the performances of baselines on 8c are relatively better than on 14c, which confirms that the combination of serving classes is reasonable and actually eases the recognition task. In general, our refined TSM (PaddlePaddle version) method achieves the second highest accuracy in all four settings except the top5 accuracy on the 14c datasets. The popular transformer-based method Vid also shows competitive performances in second place as a whole. VideoMAE V2-g almost beats all the rest of the baselines in most cases. Considering the giant backbone ViT used, it is reasonable that VideoMAE V2-g can achieve a better accuracy compared to those slim models. Note that, since the same strokes/actions done by the player facing the camera and with back towards the camera differs drastically (e.g., the blocked views and the opposite actions), we mainly focus on the facing players' actions/strokes as the recognition targets. As for the back view player's actions/strokes, the performances are even poorer for all methods. Alternatively, we report the performance of the mixed (i.e., face and back) one, the results are far below the expectation, where even the best performer, TSM, only achieves 59.46 top 1 accuracy on the 14c and 68.39 top5 accuracy on the 8c dataset. Although the top5 accuracy of all the methods appears to be promising, considering the total number of categories in \TheName{}, which is 14 or 8 accordingly, the results are trivial to some degree. In conclusion, for the action recognition task with the mainstream baseline algorithms, \TheName{} is considerably a challenging dataset and there is vast room and potential for improvement and research involvement.

For the action localization task, we summarize the performances in Figure~\ref{fig:fig1}. We selected several representative state-of-the-art action localization methods that are publicly available and retrained them on our \TheName{} dataset. We report the area under the AR-AN curve for the baselines as the measurements of the localization performance. For example, in Figure~\ref{fig:fig1}(a), we draw the AR-AN curves with a varying value of tIoU from 0.5 to 0.95 separately and one mean curve (solid blue curve) to represent the overall performance of BSN on \TheName{}. Among all the methods, TCANet+ shows the highest overall AUC score which is 47.79 in Figure~\ref{fig:fig1}(f), and also outperforms other baselines in each tIoU level. To straightforwardly compare the localization results, we visualize the located action segments from each method in Figure~\ref{fig:predict}. The top figure is on a large scale to show the predicted segments compared with the ground-truth segments in a complete 6-minute video chunk, where the bottom one zooms in a single turn for a series of consecutive actions/strokes by the player facing the camera. From such visual observations, we can find that TCANet+ can almost reproduce the short action segments from the ground-truth labeling. However, the first period of serving action is barely located and TCANet+ mis-predicts with a much shorter one instead. Compared to the action recognition task, the action localization seems more difficult on \TheName{} since the best AUC score is even in a massive gap from the average achievement of these baselines on ActivityNet~\cite{caba2015activitynet} (47.79 $\ll$ 68.25 on average from top35 solutions in the 2018 leader-board).

\begin{figure}[ht]
\begin{center}
   \includegraphics[width=0.65\textwidth]{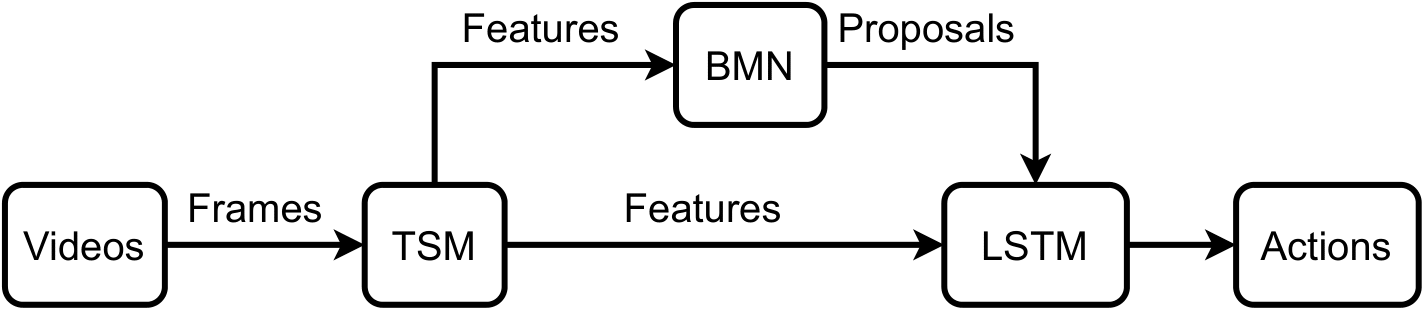}
\end{center}
   \caption{The pipeline of action detection on the \TheName{} dataset.}
\label{fig:pipeline}
\end{figure}

\noindent\textbf{Action Detection.}
Notice that the proposed \TheName{} is not only available for the action recognition and the action localization tasks. For the general action detection task, which incorporates the recognition and localization in a unified objective, the \TheName{} dataset is still challenging and remains explored in future studies. As shown in Figure~\ref{fig:pipeline}, we conduct the experiments for the action detection task on \TheName{} using a pipeline with the components of (1) TSM (PaddlePaddle variation), (2) BMN, and (3) Attention-LSTM. Specifically, we first extract the frame features using a TSM module (TSM module is pre-trained on a recognition task on the \TheName{} dataset). Once we have the well-represented features, we then feed the features into a BMN module to generate action proposals. The last step is to train an Attention-LSTM module to obtain the action class within each proposal. We further provide ablation tests in Table~\ref{tab:pipeline}. As we can observe, several optimization strategies improve the average mAP (mean average precision) of the proposed pipeline, which includes the proposal extension and data augmentation. Since the proposal generated by the BMN module sometimes cannot cover the entire action, we intend to extend the proposal duration before and after for 1 second. However, we find that not all the actions benefit from this kind of extension, where we additionally design a partial proposal extension strategy for the target actions. Although applying the above-mentioned strategies, we hardly obtain a satisfied average mAP (IoU from 0.5 to 0.9 with a step size of 0.05), where the value is only 49.74 for the best model after the fine-tuning. It makes sense with the unsatisfactory result that the performance of action detection on \TheName{} depends on the performance of both the action recognition and the action localization tasks, which have been shown as two challenging tasks in this paper. 

\begin{table}[h]
\caption{Ablation experiments of the action detection pipeline on \TheName{}. \normalfont{(``EPE/PPE'' stands for the entire/partial proposal extension. ``DA'' represents the data augmentation.)}}
    \centering
    \scalebox{0.95}{
    \begin{tabular}{c|c|c|c|c}
        \toprule
         \diagbox[width=15em, innerrightsep=5em]{Pipelines}{mAP}{IoU} & 0.5 & 0.75 & 0.95 & Average \\ 
         \hline
        TSN + BMN + LSTM & 37.71 & 16.33 & 4.10 & 18.99 \\
        TSM + BMN + LSTM & 40.39 & 22.01 & 4.94 & 20.34  \\
        TSM + BMN + LSTM + EPE & 45.14 & 22.94 & 5.89 & 22.06 \\
        TSM + BMN + LSTM + PPE & 46.03 & 25.79 & 7.17 & 24.35 \\
        TSM + DA + BMN + LSTM + PPE & 49.75 & 30.68 & 8.03 & 33.35 \\
        \bottomrule
    \end{tabular}}
    \label{tab:pipeline}
\end{table} 

\subsection{Enhancement}
Based on the above experimental results, we are eager to improve the relatively poor performance on localization tasks and provide several tips which are helpful when optimizing the baseline models. Table~\ref{tab:model-results} presents the results of several trials of optimization on BMN models, with varying configurations of parameters including the type of model, learning rate, window size, Time and Duration scales, and whether or not pre-processing, data Augmentation, fine-tuning and TTA (Test Time Augmentation~\cite{kimura2021understanding}) were carried out. The Results column represents a percentage-based performance indicator, AR-AN, of each configuration. The final enhanced model is achieved in the last row, where there are almost 4\% enhancement overall. Several key observations can be made during the optimization:

\begin{itemize}
    \item The performance of the experiments seems generally to improve as more steps were applied to the process. Comparing entries where all other parameters are held constant, runs with Preprocessing, Data Augmentation, and Fine-Tuning yield higher results than equivalent runs without these steps.
    \item Modifying the D-Scale value (Reducing it from 300 to 50) while keeping other parameters constant also leads to an improved performance.
    \item Changing the Window Size does not seem to significantly affect the results as suggested by the comparison between the first three entries. However, it affects the final result consistently since the stroke frequency in table tennis video is higher than normal sports. Considering this, it is better to choose a small window size accordingly. 
    \item Experimenting with different combinations of models appear beneficial. When BMN and TCANet+ are combined, whether by addition or multiplication (as shwon in Equation (1), the confidential scores multiplies), while keeping other parameters constant, it results in better performance than using BMN or TCANet+ individually.
    \begin{equation}
        S = S_{BMN} \times S_{TCANet+}
    \end{equation}
    \item The effectiveness of Test Time Augmentation (TTA) slightly improves the model's performance, as can be seen from comparing the final two entries.
    \item Changing the Learning Rate from CWCD (Custom Warmup Cosine Decay)~\cite{chetoui2020explainable} to CPWD (Custom Piecewise Decay)~\cite{ma2023classification} appears to improve the model performance slightly, but more runs may be needed to confirm this. We cast upon this observation and try to figure out in the future endeavours.
    \item There's a consistently increasing trend in the Results, indicating a potential correlation between the sequence of configurations and improving performance.
\end{itemize}
\begin{table}[ht]
\centering
\caption{Model Configurations and Results}
\label{tab:model-results}
\scalebox{0.60}{
\begin{tabular}{|c|c|c|c|c|c|c|c|c|c|}
\hline
\multirow{2}{*}{Model} & \multirow{2}{*}{Learning Rate} & \multirow{2}{*}{Window Size} & \multirow{2}{*}{T-Scale} & \multirow{2}{*}{D-Scale} & \multirow{2}{*}{Preprocess} & \multirow{2}{*}{Data Augmentation} & \multirow{2}{*}{Fine Tune} & \multirow{2}{*}{TTA} & \multirow{2}{*}{Results} \\
 & & & & & & & & & \\
\hline
BMN & CWCD & 8 & 300 & 300 & $\times$ & $\times$ & $\times$ & $\times$ & 43.86 \\
BMN & CWCD & 18 & 300 & 300 & $\times$ & $\times$ & $\times$ & $\times$ & 43.11 \\
BMN & CWCD & 12 & 300 & 300 & $\times$ & $\times$ & $\times$ & $\times$ & 43.25 \\
BMN & CWCD & 12 & 300 & 300 & $\times$ & $\times$ & $\checkmark$ & $\times$ & 44.10 \\
BMN & CWCD & 12 & 300 & 300 & $\times$ & $\checkmark$ & $\times$ & $\times$ & 43.64 \\
BMN & CWCD & 12 & 300 & 300 & $\times$ & $\checkmark$ & $\checkmark$ & $\times$ & 43.74 \\
BMN & CWCD & 12 & 300 & 300 & $\checkmark$ & $\checkmark$ & $\times$ & $\times$ & 44.93 \\
BMN & CWCD & 12 & 300 & 300 & $\checkmark$ & $\checkmark$ & $\checkmark$ & $\times$ & 45.76 \\
BMN & CWCD & 12 & 300 & 100 & $\checkmark$ & $\checkmark$ & $\checkmark$ & $\times$ & 46.03 \\
BMN & CWCD & 12 & 300 & 60 & $\checkmark$ & $\checkmark$ & $\checkmark$ & $\times$ & 46.69 \\
BMN & CWCD & 12 & 300 & 50 & $\checkmark$ & $\checkmark$ & $\checkmark$ & $\times$ & 46.71 \\
BMN & CWCD & 12 & 300 & 50 & $\checkmark$ & $\checkmark$ & $\checkmark$ & $\checkmark$ & 46.74 \\
BMN & CPWD & 12 & 300 & 50 & $\checkmark$ & $\checkmark$ & $\checkmark$ & $\times$ & 46.96 \\
ABMN & CPWD & 12 & 300 & 50 & $\checkmark$ & $\checkmark$ & $\checkmark$ & $\checkmark$ & 47.50 \\
BMN+TCANet+ (add) & CPWD & 12 & 300 & 50 & $\checkmark$ & $\checkmark$ & $\checkmark$ & $\times$ & 47.52 \\
BMN+TCANet+ (multiply) & CPWD & 12 & 300 & 50 & $\checkmark$ & $\checkmark$ & $\checkmark$ & $\times$ & 47.77 \\
\hline
\end{tabular}}
\end{table}

\subsection{Remarks and Summary}
In this section, we mainly target the 2D video action recognition algorithms considering the following sakes,
\begin{itemize}[leftmargin=*]
    \item Traditional 3D/4D video action recognition algorithms such as I3D~\cite{carreira2017quo}, R3D~\cite{hara2018can}, S3D~\cite{xie2018rethinking}, and Non-local~\cite{wang2018non} are widely acknowledged to be inefficient in computation and optimization compared to the 2D algorithms (i.e., it is barely affordable and scalable in some real-world applications), even though the 3D/4D algorithms generally have a more robust performance~\cite{zhu2020comprehensive} on video recognition tasks.
    \item Analyzing table tennis broadcasting video is not only limited to academic purposes, which includes the research on dense, fast-moving, and noisy (e.g., multiple action sources from players) action recognition and localization, it also benefits table tennis sports affairs, for example, real-time competition analysis, and actions/events summarizing on stream broadcasting videos. Thus, effectiveness and efficiency are of equal importance for the solutions in industrial practices, where the 2D and lightweight video action recognition algorithms have received a great deal of industrial and research attention in recent years.
    \item We also include one 3D action recognition algorithm -- SlowFast~\cite{feichtenhofer2019slowfast}. Compared to the traditional 3D networks, SlowFast does not heavily rely on the concatenation of 3D CNNs, where its fast pathway could be very lightweight by reducing its channel capacity, so as to largely improve the overall efficiency. Furthermore, SlowFast is proven to be competitive with other full-size 3D recognition algorithms on several benchmark datasets~\cite{zhu2020comprehensive}.
    \item Biases can originate from various parts of the data collection and annotation process, such as game selection, player selection, or even specific play styles that might dominate the dataset. In future iterations, we plan to diversify our dataset sources to minimize these types of bias.
    \item Table tennis is a versatile sport with players exhibiting a wide array of unique styles and techniques which can directly impact the recognition and classification. We aim to augment our dataset to account for these stylistic variations including different types of grips, serves, strokes, player orientation, and attack or defense playstyles.
    \item Some other limiting factors could be the quality of video recording, the varying level of professional play, or obstructions in field view. Additionally, crowd noise and other audio elements might impact audio-based action recognition. We'll address these in the revisions and outline how different kinds of noise and visual obstruction were dealt with.
\end{itemize}

From the above issues, we can conclude that the \TheName{} is still a very challenging dataset no matter the action recognition and the action localization tasks. Compared to the current release sports datasets, the representative baselines hardly adapt to the proposed \TheName{} dataset and the performance is far lower than our expectation. Thus, the \TheName{} dataset provides a good opportunity for researchers to further explore the solution in a domain of fast-moving and dense action detection. Note that, the \TheName{} dataset is not limited to the action recognition and localization tasks only, where it is also potential to be used for video segmentation and summarizing. Actually, we are on the way to exploring more possibilities of \TheName{} and will update the released version to include more use cases. Due to the limited space, we will release the implementation details for all the baselines in our GitHub repository with the dataset link once the paper gets published.

\section{Conclusion}
In this paper, we present a new large-scale benchmark \TheName{}, which contains currently the largest publicly available table tennis action detection dataset to the best of our knowledge. We have evaluated several state-of-the-art algorithms on the introduced dataset. Experimental results show that existing approaches are heavily challenged by \textbf{fast-moving} and \textbf{dense} actions and uneven action distribution in video chunks. In the future, we hope that \TheName{} could become a meaningful benchmark for fast-moving and dense action detection, especially in the video analysis area. Through a comprehensive introduction and analysis, we intend to help the follow-up users and researchers to better acknowledge the characteristics and the statistics of \TheName{}, so as to fully utilize the dataset in action recognition and action localization tasks. We believe the \TheName{} dataset will make an early attempt in driving the sports video analysis industry to be more intelligent and effective.

\bibliographystyle{ACM-Reference-Format}
\bibliography{sample-base}


\begin{thebibliography}{74}


\ifx \showCODEN    \undefined \def \showCODEN     #1{\unskip}     \fi
\ifx \showDOI      \undefined \def \showDOI       #1{#1}\fi
\ifx \showISBNx    \undefined \def \showISBNx     #1{\unskip}     \fi
\ifx \showISBNxiii \undefined \def \showISBNxiii  #1{\unskip}     \fi
\ifx \showISSN     \undefined \def \showISSN      #1{\unskip}     \fi
\ifx \showLCCN     \undefined \def \showLCCN      #1{\unskip}     \fi
\ifx \shownote     \undefined \def \shownote      #1{#1}          \fi
\ifx \showarticletitle \undefined \def \showarticletitle #1{#1}   \fi
\ifx \showURL      \undefined \def \showURL       {\relax}        \fi
\providecommand\bibfield[2]{#2}
\providecommand\bibinfo[2]{#2}
\providecommand\natexlab[1]{#1}
\providecommand\showeprint[2][]{arXiv:#2}

\bibitem[Abu-El-Haija et~al\mbox{.}(2016)]%
        {abu2016youtube}
\bibfield{author}{\bibinfo{person}{Sami Abu-El-Haija}, \bibinfo{person}{Nisarg
  Kothari}, \bibinfo{person}{Joonseok Lee}, \bibinfo{person}{Paul Natsev},
  \bibinfo{person}{George Toderici}, \bibinfo{person}{Balakrishnan
  Varadarajan}, {and} \bibinfo{person}{Sudheendra Vijayanarasimhan}.}
  \bibinfo{year}{2016}\natexlab{}.
\newblock \showarticletitle{Youtube-8m: A large-scale video classification
  benchmark}.
\newblock \bibinfo{journal}{\emph{arXiv preprint arXiv:1609.08675}}
  (\bibinfo{year}{2016}).
\newblock


\bibitem[Arnab et~al\mbox{.}(2021)]%
        {arnab2021vivit}
\bibfield{author}{\bibinfo{person}{Anurag Arnab}, \bibinfo{person}{Mostafa
  Dehghani}, \bibinfo{person}{Georg Heigold}, \bibinfo{person}{Chen Sun},
  \bibinfo{person}{Mario Lu{\v{c}}i{\'c}}, {and} \bibinfo{person}{Cordelia
  Schmid}.} \bibinfo{year}{2021}\natexlab{}.
\newblock \showarticletitle{Vivit: A video vision transformer}. In
  \bibinfo{booktitle}{\emph{Proceedings of the IEEE/CVF International
  Conference on Computer Vision}}. \bibinfo{pages}{6836--6846}.
\newblock


\bibitem[Bao and Yao(2021)]%
        {bao2021dynamic}
\bibfield{author}{\bibinfo{person}{Hongshu Bao} {and} \bibinfo{person}{Xiang
  Yao}.} \bibinfo{year}{2021}\natexlab{}.
\newblock \showarticletitle{RETRACTED: Dynamic 3D image simulation of
  basketball movement based on embedded system and computer vision}.
\newblock \bibinfo{journal}{\emph{Microprocessors and Microsystems}}
  \bibinfo{volume}{81} (\bibinfo{year}{2021}), \bibinfo{pages}{103655}.
\newblock
\showISSN{0141-9331}
\urldef\tempurl%
\url{https://doi.org/10.1016/j.micpro.2020.103655}
\showDOI{\tempurl}


\bibitem[Bertasius et~al\mbox{.}(2017)]%
        {bertasius2017baller}
\bibfield{author}{\bibinfo{person}{Gedas Bertasius}, \bibinfo{person}{Hyun
  Soo~Park}, \bibinfo{person}{Stella~X Yu}, {and} \bibinfo{person}{Jianbo
  Shi}.} \bibinfo{year}{2017}\natexlab{}.
\newblock \showarticletitle{Am I a baller? basketball performance assessment
  from first-person videos}. In \bibinfo{booktitle}{\emph{Proceedings of the
  IEEE international conference on computer vision}}.
  \bibinfo{pages}{2177--2185}.
\newblock


\bibitem[Bertasius et~al\mbox{.}(2021)]%
        {bertasius2021space}
\bibfield{author}{\bibinfo{person}{Gedas Bertasius}, \bibinfo{person}{Heng
  Wang}, {and} \bibinfo{person}{Lorenzo Torresani}.}
  \bibinfo{year}{2021}\natexlab{}.
\newblock \showarticletitle{Is Space-Time Attention All You Need for Video
  Understanding?}. In \bibinfo{booktitle}{\emph{International Conference on
  Machine Learning}}. PMLR, \bibinfo{pages}{813--824}.
\newblock


\bibitem[Bian et~al\mbox{.}(2022)]%
        {bian2022machine}
\bibfield{author}{\bibinfo{person}{Jiang Bian}, \bibinfo{person}{Abdullah
  Al~Arafat}, \bibinfo{person}{Haoyi Xiong}, \bibinfo{person}{Jing Li},
  \bibinfo{person}{Li Li}, \bibinfo{person}{Hongyang Chen},
  \bibinfo{person}{Jun Wang}, \bibinfo{person}{Dejing Dou}, {and}
  \bibinfo{person}{Zhishan Guo}.} \bibinfo{year}{2022}\natexlab{}.
\newblock \showarticletitle{Machine learning in real-time internet of things
  (iot) systems: A survey}.
\newblock \bibinfo{journal}{\emph{IEEE Internet of Things Journal}}
  \bibinfo{volume}{9}, \bibinfo{number}{11} (\bibinfo{year}{2022}),
  \bibinfo{pages}{8364--8386}.
\newblock


\bibitem[Caba~Heilbron et~al\mbox{.}(2015)]%
        {caba2015activitynet}
\bibfield{author}{\bibinfo{person}{Fabian Caba~Heilbron},
  \bibinfo{person}{Victor Escorcia}, \bibinfo{person}{Bernard Ghanem}, {and}
  \bibinfo{person}{Juan Carlos~Niebles}.} \bibinfo{year}{2015}\natexlab{}.
\newblock \showarticletitle{Activitynet: A large-scale video benchmark for
  human activity understanding}. In \bibinfo{booktitle}{\emph{Proceedings of
  the IEEE/CVF conference on computer vision and pattern recognition}}.
  \bibinfo{pages}{961--970}.
\newblock


\bibitem[Cai et~al\mbox{.}(2023)]%
        {cai2022heterogeneous}
\bibfield{author}{\bibinfo{person}{Desheng Cai}, \bibinfo{person}{Shengsheng
  Qian}, \bibinfo{person}{Quan Fang}, \bibinfo{person}{Jun Hu},
  \bibinfo{person}{Wenkui Ding}, {and} \bibinfo{person}{Changsheng Xu}.}
  \bibinfo{year}{2023}\natexlab{}.
\newblock \showarticletitle{Heterogeneous Graph Contrastive Learning Network
  for Personalized Micro-Video Recommendation}.
\newblock \bibinfo{journal}{\emph{IEEE Transactions on Multimedia}}
  \bibinfo{volume}{25} (\bibinfo{year}{2023}), \bibinfo{pages}{2761--2773}.
\newblock
\urldef\tempurl%
\url{https://doi.org/10.1109/TMM.2022.3151026}
\showDOI{\tempurl}


\bibitem[Carreira and Zisserman(2017)]%
        {carreira2017quo}
\bibfield{author}{\bibinfo{person}{Joao Carreira} {and} \bibinfo{person}{Andrew
  Zisserman}.} \bibinfo{year}{2017}\natexlab{}.
\newblock \showarticletitle{Quo vadis, action recognition? a new model and the
  kinetics dataset}. In \bibinfo{booktitle}{\emph{proceedings of the IEEE
  Conference on Computer Vision and Pattern Recognition}}.
  \bibinfo{pages}{6299--6308}.
\newblock


\bibitem[Chetoui and Akhloufi(2020)]%
        {chetoui2020explainable}
\bibfield{author}{\bibinfo{person}{Mohamed Chetoui} {and}
  \bibinfo{person}{Moulay~A Akhloufi}.} \bibinfo{year}{2020}\natexlab{}.
\newblock \showarticletitle{Explainable end-to-end deep learning for diabetic
  retinopathy detection across multiple datasets}.
\newblock \bibinfo{journal}{\emph{Journal of Medical Imaging}}
  \bibinfo{volume}{7}, \bibinfo{number}{4} (\bibinfo{year}{2020}),
  \bibinfo{pages}{044503--044503}.
\newblock


\bibitem[Cubuk et~al\mbox{.}(2018)]%
        {cubuk2018autoaugment}
\bibfield{author}{\bibinfo{person}{Ekin~D Cubuk}, \bibinfo{person}{Barret
  Zoph}, \bibinfo{person}{Dandelion Mane}, \bibinfo{person}{Vijay Vasudevan},
  {and} \bibinfo{person}{Quoc~V Le}.} \bibinfo{year}{2018}\natexlab{}.
\newblock \showarticletitle{Autoaugment: Learning augmentation policies from
  data}.
\newblock \bibinfo{journal}{\emph{arXiv preprint arXiv:1805.09501}}
  (\bibinfo{year}{2018}).
\newblock


\bibitem[Dai et~al\mbox{.}(2017)]%
        {dai2017temporal}
\bibfield{author}{\bibinfo{person}{Xiyang Dai}, \bibinfo{person}{Bharat Singh},
  \bibinfo{person}{Guyue Zhang}, \bibinfo{person}{Larry~S Davis}, {and}
  \bibinfo{person}{Yan Qiu~Chen}.} \bibinfo{year}{2017}\natexlab{}.
\newblock \showarticletitle{Temporal context network for activity localization
  in videos}. In \bibinfo{booktitle}{\emph{Proceedings of the IEEE
  International Conference on Computer Vision}}. \bibinfo{pages}{5793--5802}.
\newblock


\bibitem[Dalal and Triggs(2005)]%
        {dalal2005histograms}
\bibfield{author}{\bibinfo{person}{Navneet Dalal} {and} \bibinfo{person}{Bill
  Triggs}.} \bibinfo{year}{2005}\natexlab{}.
\newblock \showarticletitle{Histograms of oriented gradients for human
  detection}. In \bibinfo{booktitle}{\emph{2005 IEEE computer society
  conference on computer vision and pattern recognition (CVPR'05)}},
  Vol.~\bibinfo{volume}{1}. Ieee, \bibinfo{pages}{886--893}.
\newblock


\bibitem[Deliege et~al\mbox{.}(2021)]%
        {deliege2021soccernet}
\bibfield{author}{\bibinfo{person}{Adrien Deliege}, \bibinfo{person}{Anthony
  Cioppa}, \bibinfo{person}{Silvio Giancola}, \bibinfo{person}{Meisam~J
  Seikavandi}, \bibinfo{person}{Jacob~V Dueholm}, \bibinfo{person}{Kamal
  Nasrollahi}, \bibinfo{person}{Bernard Ghanem}, \bibinfo{person}{Thomas~B
  Moeslund}, {and} \bibinfo{person}{Marc Van~Droogenbroeck}.}
  \bibinfo{year}{2021}\natexlab{}.
\newblock \showarticletitle{Soccernet-v2: A dataset and benchmarks for holistic
  understanding of broadcast soccer videos}. In
  \bibinfo{booktitle}{\emph{Proceedings of the IEEE/CVF Conference on Computer
  Vision and Pattern Recognition}}. \bibinfo{pages}{4508--4519}.
\newblock


\bibitem[Dong et~al\mbox{.}(2022)]%
        {dong2021dual}
\bibfield{author}{\bibinfo{person}{Jianfeng Dong}, \bibinfo{person}{Xirong Li},
  \bibinfo{person}{Chaoxi Xu}, \bibinfo{person}{Xun Yang},
  \bibinfo{person}{Gang Yang}, \bibinfo{person}{Xun Wang}, {and}
  \bibinfo{person}{Meng Wang}.} \bibinfo{year}{2022}\natexlab{}.
\newblock \showarticletitle{Dual Encoding for Video Retrieval by Text}.
\newblock \bibinfo{journal}{\emph{IEEE Trans. Pattern Anal. Mach. Intell.}}
  \bibinfo{volume}{44}, \bibinfo{number}{8} (\bibinfo{date}{aug}
  \bibinfo{year}{2022}), \bibinfo{pages}{4065–4080}.
\newblock
\showISSN{0162-8828}
\urldef\tempurl%
\url{https://doi.org/10.1109/TPAMI.2021.3059295}
\showDOI{\tempurl}


\bibitem[Faulkner and Dick(2017)]%
        {faulkner2017tenniset}
\bibfield{author}{\bibinfo{person}{Hayden Faulkner} {and}
  \bibinfo{person}{Anthony Dick}.} \bibinfo{year}{2017}\natexlab{}.
\newblock \showarticletitle{TenniSet: A Dataset for Dense Fine-Grained Event
  Recognition, Localisation and Description}. In \bibinfo{booktitle}{\emph{2017
  International Conference on Digital Image Computing: Techniques and
  Applications (DICTA)}}. IEEE, \bibinfo{pages}{1--8}.
\newblock


\bibitem[Feichtenhofer et~al\mbox{.}(2019)]%
        {feichtenhofer2019slowfast}
\bibfield{author}{\bibinfo{person}{Christoph Feichtenhofer},
  \bibinfo{person}{Haoqi Fan}, \bibinfo{person}{Jitendra Malik}, {and}
  \bibinfo{person}{Kaiming He}.} \bibinfo{year}{2019}\natexlab{}.
\newblock \showarticletitle{Slowfast networks for video recognition}. In
  \bibinfo{booktitle}{\emph{Proceedings of the IEEE/CVF international
  conference on computer vision}}. \bibinfo{pages}{6202--6211}.
\newblock


\bibitem[Gabeur et~al\mbox{.}(2020)]%
        {gabeur2020multi}
\bibfield{author}{\bibinfo{person}{Valentin Gabeur}, \bibinfo{person}{Chen
  Sun}, \bibinfo{person}{Karteek Alahari}, {and} \bibinfo{person}{Cordelia
  Schmid}.} \bibinfo{year}{2020}\natexlab{}.
\newblock \showarticletitle{Multi-modal transformer for video retrieval}. In
  \bibinfo{booktitle}{\emph{European Conference on Computer Vision}}. Springer,
  \bibinfo{pages}{214--229}.
\newblock


\bibitem[Goyal et~al\mbox{.}(2017)]%
        {goyal2017something}
\bibfield{author}{\bibinfo{person}{Raghav Goyal}, \bibinfo{person}{Samira
  Ebrahimi~Kahou}, \bibinfo{person}{Vincent Michalski}, \bibinfo{person}{Joanna
  Materzynska}, \bibinfo{person}{Susanne Westphal}, \bibinfo{person}{Heuna
  Kim}, \bibinfo{person}{Valentin Haenel}, \bibinfo{person}{Ingo Fruend},
  \bibinfo{person}{Peter Yianilos}, \bibinfo{person}{Moritz Mueller-Freitag},
  {et~al\mbox{.}}} \bibinfo{year}{2017}\natexlab{}.
\newblock \showarticletitle{The" something something" video database for
  learning and evaluating visual common sense}. In
  \bibinfo{booktitle}{\emph{Proceedings of the IEEE international conference on
  computer vision}}. \bibinfo{pages}{5842--5850}.
\newblock


\bibitem[Gu et~al\mbox{.}(2018)]%
        {gu2018ava}
\bibfield{author}{\bibinfo{person}{Chunhui Gu}, \bibinfo{person}{Chen Sun},
  \bibinfo{person}{David~A Ross}, \bibinfo{person}{Carl Vondrick},
  \bibinfo{person}{Caroline Pantofaru}, \bibinfo{person}{Yeqing Li},
  \bibinfo{person}{Sudheendra Vijayanarasimhan}, \bibinfo{person}{George
  Toderici}, \bibinfo{person}{Susanna Ricco}, \bibinfo{person}{Rahul
  Sukthankar}, {et~al\mbox{.}}} \bibinfo{year}{2018}\natexlab{}.
\newblock \showarticletitle{Ava: A video dataset of spatio-temporally localized
  atomic visual actions}. In \bibinfo{booktitle}{\emph{Proceedings of the IEEE
  Conference on Computer Vision and Pattern Recognition}}.
  \bibinfo{pages}{6047--6056}.
\newblock


\bibitem[Hara et~al\mbox{.}(2018)]%
        {hara2018can}
\bibfield{author}{\bibinfo{person}{Kensho Hara}, \bibinfo{person}{Hirokatsu
  Kataoka}, {and} \bibinfo{person}{Yutaka Satoh}.}
  \bibinfo{year}{2018}\natexlab{}.
\newblock \showarticletitle{Can spatiotemporal 3d cnns retrace the history of
  2d cnns and imagenet?}. In \bibinfo{booktitle}{\emph{Proceedings of the IEEE
  conference on Computer Vision and Pattern Recognition}}.
  \bibinfo{pages}{6546--6555}.
\newblock


\bibitem[Ibrahim et~al\mbox{.}(2016)]%
        {ibrahim2016hierarchical}
\bibfield{author}{\bibinfo{person}{Mostafa~S Ibrahim},
  \bibinfo{person}{Srikanth Muralidharan}, \bibinfo{person}{Zhiwei Deng},
  \bibinfo{person}{Arash Vahdat}, {and} \bibinfo{person}{Greg Mori}.}
  \bibinfo{year}{2016}\natexlab{}.
\newblock \showarticletitle{A hierarchical deep temporal model for group
  activity recognition}. In \bibinfo{booktitle}{\emph{Proceedings of the IEEE
  conference on computer vision and pattern recognition}}.
  \bibinfo{pages}{1971--1980}.
\newblock


\bibitem[Jiang et~al\mbox{.}(2020)]%
        {jiang2020soccerdb}
\bibfield{author}{\bibinfo{person}{Yudong Jiang}, \bibinfo{person}{Kaixu Cui},
  \bibinfo{person}{Leilei Chen}, \bibinfo{person}{Canjin Wang}, {and}
  \bibinfo{person}{Changliang Xu}.} \bibinfo{year}{2020}\natexlab{}.
\newblock \showarticletitle{Soccerdb: A large-scale database for comprehensive
  video understanding}. In \bibinfo{booktitle}{\emph{Proceedings of the 3rd
  International Workshop on Multimedia Content Analysis in Sports}}.
  \bibinfo{pages}{1--8}.
\newblock


\bibitem[Karpathy et~al\mbox{.}(2014)]%
        {karpathy2014large}
\bibfield{author}{\bibinfo{person}{Andrej Karpathy}, \bibinfo{person}{George
  Toderici}, \bibinfo{person}{Sanketh Shetty}, \bibinfo{person}{Thomas Leung},
  \bibinfo{person}{Rahul Sukthankar}, {and} \bibinfo{person}{Li Fei-Fei}.}
  \bibinfo{year}{2014}\natexlab{}.
\newblock \showarticletitle{Large-scale video classification with convolutional
  neural networks}. In \bibinfo{booktitle}{\emph{Proceedings of the IEEE
  conference on Computer Vision and Pattern Recognition}}.
  \bibinfo{pages}{1725--1732}.
\newblock


\bibitem[Kimura(2021)]%
        {kimura2021understanding}
\bibfield{author}{\bibinfo{person}{Masanari Kimura}.}
  \bibinfo{year}{2021}\natexlab{}.
\newblock \showarticletitle{Understanding test-time augmentation}. In
  \bibinfo{booktitle}{\emph{International Conference on Neural Information
  Processing}}. Springer, \bibinfo{pages}{558--569}.
\newblock


\bibitem[Kondratyuk et~al\mbox{.}(2021)]%
        {kondratyuk2021movinets}
\bibfield{author}{\bibinfo{person}{Dan Kondratyuk}, \bibinfo{person}{Liangzhe
  Yuan}, \bibinfo{person}{Yandong Li}, \bibinfo{person}{Li Zhang},
  \bibinfo{person}{Mingxing Tan}, \bibinfo{person}{Matthew Brown}, {and}
  \bibinfo{person}{Boqing Gong}.} \bibinfo{year}{2021}\natexlab{}.
\newblock \showarticletitle{Movinets: Mobile video networks for efficient video
  recognition}. In \bibinfo{booktitle}{\emph{Proceedings of the IEEE/CVF
  Conference on Computer Vision and Pattern Recognition}}.
  \bibinfo{pages}{16020--16030}.
\newblock


\bibitem[Koshkina et~al\mbox{.}(2021)]%
        {koshkina2021contrastive}
\bibfield{author}{\bibinfo{person}{Maria Koshkina}, \bibinfo{person}{Hemanth
  Pidaparthy}, {and} \bibinfo{person}{James~H Elder}.}
  \bibinfo{year}{2021}\natexlab{}.
\newblock \showarticletitle{Contrastive learning for sports video: Unsupervised
  player classification}. In \bibinfo{booktitle}{\emph{Proceedings of the
  IEEE/CVF Conference on Computer Vision and Pattern Recognition}}.
  \bibinfo{pages}{4528--4536}.
\newblock


\bibitem[Kulkarni and Shenoy(2021)]%
        {kulkarni2021table}
\bibfield{author}{\bibinfo{person}{Kaustubh~Milind Kulkarni} {and}
  \bibinfo{person}{Sucheth Shenoy}.} \bibinfo{year}{2021}\natexlab{}.
\newblock \showarticletitle{Table tennis stroke recognition using
  two-dimensional human pose estimation}. In
  \bibinfo{booktitle}{\emph{Proceedings of the IEEE/CVF Conference on Computer
  Vision and Pattern Recognition}}. \bibinfo{pages}{4576--4584}.
\newblock


\bibitem[Lin et~al\mbox{.}(2019a)]%
        {lin2019tsm}
\bibfield{author}{\bibinfo{person}{Ji Lin}, \bibinfo{person}{Chuang Gan}, {and}
  \bibinfo{person}{Song Han}.} \bibinfo{year}{2019}\natexlab{a}.
\newblock \showarticletitle{Tsm: Temporal shift module for efficient video
  understanding}. In \bibinfo{booktitle}{\emph{Proceedings of the IEEE/CVF
  International Conference on Computer Vision}}. \bibinfo{pages}{7083--7093}.
\newblock


\bibitem[Lin et~al\mbox{.}(2019b)]%
        {lin2019bmn}
\bibfield{author}{\bibinfo{person}{Tianwei Lin}, \bibinfo{person}{Xiao Liu},
  \bibinfo{person}{Xin Li}, \bibinfo{person}{Errui Ding}, {and}
  \bibinfo{person}{Shilei Wen}.} \bibinfo{year}{2019}\natexlab{b}.
\newblock \showarticletitle{Bmn: Boundary-matching network for temporal action
  proposal generation}. In \bibinfo{booktitle}{\emph{Proceedings of the
  IEEE/CVF International Conference on Computer Vision}}.
  \bibinfo{pages}{3889--3898}.
\newblock


\bibitem[Lin et~al\mbox{.}(2018)]%
        {lin2018bsn}
\bibfield{author}{\bibinfo{person}{Tianwei Lin}, \bibinfo{person}{Xu Zhao},
  \bibinfo{person}{Haisheng Su}, \bibinfo{person}{Chongjing Wang}, {and}
  \bibinfo{person}{Ming Yang}.} \bibinfo{year}{2018}\natexlab{}.
\newblock \showarticletitle{Bsn: Boundary sensitive network for temporal action
  proposal generation}. In \bibinfo{booktitle}{\emph{Proceedings of the
  European conference on computer vision (ECCV)}}. \bibinfo{pages}{3--19}.
\newblock


\bibitem[Liu et~al\mbox{.}(2017)]%
        {liu2017deep}
\bibfield{author}{\bibinfo{person}{Wu Liu}, \bibinfo{person}{Chenggang~Clarence
  Yan}, \bibinfo{person}{Jiangyu Liu}, {and} \bibinfo{person}{Huadong Ma}.}
  \bibinfo{year}{2017}\natexlab{}.
\newblock \showarticletitle{Deep learning based basketball video analysis for
  intelligent arena application}.
\newblock \bibinfo{journal}{\emph{Multimedia Tools and Applications}}
  \bibinfo{volume}{76}, \bibinfo{number}{23} (\bibinfo{year}{2017}),
  \bibinfo{pages}{24983--25001}.
\newblock


\bibitem[Liu et~al\mbox{.}(2019a)]%
        {liu2019use}
\bibfield{author}{\bibinfo{person}{Yang Liu}, \bibinfo{person}{Samuel Albanie},
  \bibinfo{person}{Arsha Nagrani}, {and} \bibinfo{person}{Andrew Zisserman}.}
  \bibinfo{year}{2019}\natexlab{a}.
\newblock \showarticletitle{Use what you have: Video retrieval using
  representations from collaborative experts}.
\newblock \bibinfo{journal}{\emph{arXiv preprint arXiv:1907.13487}}
  (\bibinfo{year}{2019}).
\newblock


\bibitem[Liu et~al\mbox{.}(2019b)]%
        {liu2019building}
\bibfield{author}{\bibinfo{person}{Yang Liu}, \bibinfo{person}{Cheng Lyu},
  \bibinfo{person}{Zhiyuan Liu}, {and} \bibinfo{person}{Dacheng Tao}.}
  \bibinfo{year}{2019}\natexlab{b}.
\newblock \showarticletitle{Building effective short video recommendation}. In
  \bibinfo{booktitle}{\emph{2019 IEEE International Conference on Multimedia \&
  Expo Workshops (ICMEW)}}. IEEE, \bibinfo{pages}{651--656}.
\newblock


\bibitem[Liu et~al\mbox{.}(2021)]%
        {liu2021swin}
\bibfield{author}{\bibinfo{person}{Ze Liu}, \bibinfo{person}{Yutong Lin},
  \bibinfo{person}{Yue Cao}, \bibinfo{person}{Han Hu}, \bibinfo{person}{Yixuan
  Wei}, \bibinfo{person}{Zheng Zhang}, \bibinfo{person}{Stephen Lin}, {and}
  \bibinfo{person}{Baining Guo}.} \bibinfo{year}{2021}\natexlab{}.
\newblock \showarticletitle{Swin transformer: Hierarchical vision transformer
  using shifted windows}. In \bibinfo{booktitle}{\emph{Proceedings of the
  IEEE/CVF International Conference on Computer Vision}}.
  \bibinfo{pages}{10012--10022}.
\newblock


\bibitem[Liu et~al\mbox{.}(2022)]%
        {liu_2021}
\bibfield{author}{\bibinfo{person}{Ze Liu}, \bibinfo{person}{Jia Ning},
  \bibinfo{person}{Yue Cao}, \bibinfo{person}{Yixuan Wei},
  \bibinfo{person}{Zheng Zhang}, \bibinfo{person}{Stephen Lin}, {and}
  \bibinfo{person}{Han Hu}.} \bibinfo{year}{2022}\natexlab{}.
\newblock \showarticletitle{Video swin transformer}. In
  \bibinfo{booktitle}{\emph{Proceedings of the IEEE/CVF conference on computer
  vision and pattern recognition}}. \bibinfo{pages}{3202--3211}.
\newblock


\bibitem[Long et~al\mbox{.}(2018)]%
        {long2018attention}
\bibfield{author}{\bibinfo{person}{Xiang Long}, \bibinfo{person}{Chuang Gan},
  \bibinfo{person}{Gerard De~Melo}, \bibinfo{person}{Jiajun Wu},
  \bibinfo{person}{Xiao Liu}, {and} \bibinfo{person}{Shilei Wen}.}
  \bibinfo{year}{2018}\natexlab{}.
\newblock \showarticletitle{Attention clusters: Purely attention based local
  feature integration for video classification}. In
  \bibinfo{booktitle}{\emph{Proceedings of the IEEE Conference on Computer
  Vision and Pattern Recognition}}. \bibinfo{pages}{7834--7843}.
\newblock


\bibitem[Ma et~al\mbox{.}(2023)]%
        {ma2023classification}
\bibfield{author}{\bibinfo{person}{Xiang Ma}, \bibinfo{person}{Yonglei Li},
  \bibinfo{person}{Lipengcheng Wan}, \bibinfo{person}{Zexin Xu},
  \bibinfo{person}{Jiannong Song}, {and} \bibinfo{person}{Jinqiu Huang}.}
  \bibinfo{year}{2023}\natexlab{}.
\newblock \showarticletitle{Classification of seed corn ears based on custom
  lightweight convolutional neural network and improved training strategies}.
\newblock \bibinfo{journal}{\emph{Engineering Applications of Artificial
  Intelligence}}  \bibinfo{volume}{120} (\bibinfo{year}{2023}),
  \bibinfo{pages}{105936}.
\newblock


\bibitem[Ma et~al\mbox{.}(2019)]%
        {ma2019paddlepaddle}
\bibfield{author}{\bibinfo{person}{Yanjun Ma}, \bibinfo{person}{Dianhai Yu},
  \bibinfo{person}{Tian Wu}, {and} \bibinfo{person}{Haifeng Wang}.}
  \bibinfo{year}{2019}\natexlab{}.
\newblock \showarticletitle{PaddlePaddle: An open-source deep learning platform
  from industrial practice}.
\newblock \bibinfo{journal}{\emph{Frontiers of Data and Domputing}}
  \bibinfo{volume}{1}, \bibinfo{number}{1} (\bibinfo{year}{2019}),
  \bibinfo{pages}{105--115}.
\newblock


\bibitem[Martin et~al\mbox{.}(2018)]%
        {martin2018sport}
\bibfield{author}{\bibinfo{person}{Pierre-Etienne Martin},
  \bibinfo{person}{Jenny Benois-Pineau}, \bibinfo{person}{Renaud P{\'e}teri},
  {and} \bibinfo{person}{Julien Morlier}.} \bibinfo{year}{2018}\natexlab{}.
\newblock \showarticletitle{Sport action recognition with siamese
  spatio-temporal cnns: Application to table tennis}. In
  \bibinfo{booktitle}{\emph{2018 International Conference on Content-Based
  Multimedia Indexing (CBMI)}}. IEEE, \bibinfo{pages}{1--6}.
\newblock


\bibitem[Martin et~al\mbox{.}(2021)]%
        {martin2021three}
\bibfield{author}{\bibinfo{person}{Pierre-Etienne Martin},
  \bibinfo{person}{Jenny Benois-Pineau}, \bibinfo{person}{Renaud P{\'e}teri},
  {and} \bibinfo{person}{Julien Morlier}.} \bibinfo{year}{2021}\natexlab{}.
\newblock \showarticletitle{Three-Stream 3D/1D CNN for Fine-Grained Action
  Classification and Segmentation in Table Tennis}. In
  \bibinfo{booktitle}{\emph{Proceedings of the 4th International Workshop on
  Multimedia Content Analysis in Sports}}. \bibinfo{pages}{35--41}.
\newblock


\bibitem[Nawhal and Mori(2021)]%
        {nawhal2021activity}
\bibfield{author}{\bibinfo{person}{Megha Nawhal} {and} \bibinfo{person}{Greg
  Mori}.} \bibinfo{year}{2021}\natexlab{}.
\newblock \showarticletitle{Activity graph transformer for temporal action
  localization}.
\newblock \bibinfo{journal}{\emph{arXiv preprint arXiv:2101.08540}}
  (\bibinfo{year}{2021}).
\newblock


\bibitem[Niebles et~al\mbox{.}(2010)]%
        {niebles2010modeling}
\bibfield{author}{\bibinfo{person}{Juan~Carlos Niebles},
  \bibinfo{person}{Chih-Wei Chen}, {and} \bibinfo{person}{Li Fei-Fei}.}
  \bibinfo{year}{2010}\natexlab{}.
\newblock \showarticletitle{Modeling temporal structure of decomposable motion
  segments for activity classification}. In \bibinfo{booktitle}{\emph{European
  conference on computer vision}}. Springer, \bibinfo{pages}{392--405}.
\newblock


\bibitem[Olympedia(2012)]%
        {wufei}
\bibfield{author}{\bibinfo{person}{Olympedia}.}
  \bibinfo{year}{2012}\natexlab{}.
\newblock \bibinfo{booktitle}{\emph{Referee Biographical Information}}.
\newblock
\urldef\tempurl%
\url{http://www.olympedia.org/athletes/5004924}
\showURL{%
\tempurl}


\bibitem[Qing et~al\mbox{.}(2021)]%
        {qing2021temporal}
\bibfield{author}{\bibinfo{person}{Zhiwu Qing}, \bibinfo{person}{Haisheng Su},
  \bibinfo{person}{Weihao Gan}, \bibinfo{person}{Dongliang Wang},
  \bibinfo{person}{Wei Wu}, \bibinfo{person}{Xiang Wang}, \bibinfo{person}{Yu
  Qiao}, \bibinfo{person}{Junjie Yan}, \bibinfo{person}{Changxin Gao}, {and}
  \bibinfo{person}{Nong Sang}.} \bibinfo{year}{2021}\natexlab{}.
\newblock \showarticletitle{Temporal context aggregation network for temporal
  action proposal refinement}. In \bibinfo{booktitle}{\emph{Proceedings of the
  IEEE/CVF Conference on Computer Vision and Pattern Recognition}}.
  \bibinfo{pages}{485--494}.
\newblock


\bibitem[Ren et~al\mbox{.}(2015)]%
        {ren2015faster}
\bibfield{author}{\bibinfo{person}{Shaoqing Ren}, \bibinfo{person}{Kaiming He},
  \bibinfo{person}{Ross Girshick}, {and} \bibinfo{person}{Jian Sun}.}
  \bibinfo{year}{2015}\natexlab{}.
\newblock \showarticletitle{Faster r-cnn: Towards real-time object detection
  with region proposal networks}.
\newblock \bibinfo{journal}{\emph{Advances in neural information processing
  systems}}  \bibinfo{volume}{28} (\bibinfo{year}{2015}).
\newblock


\bibitem[Shao et~al\mbox{.}(2020)]%
        {shao2020finegym}
\bibfield{author}{\bibinfo{person}{Dian Shao}, \bibinfo{person}{Yue Zhao},
  \bibinfo{person}{Bo Dai}, {and} \bibinfo{person}{Dahua Lin}.}
  \bibinfo{year}{2020}\natexlab{}.
\newblock \showarticletitle{Finegym: A hierarchical video dataset for
  fine-grained action understanding}. In \bibinfo{booktitle}{\emph{Proceedings
  of the IEEE/CVF conference on computer vision and pattern recognition}}.
  \bibinfo{pages}{2616--2625}.
\newblock


\bibitem[Shou et~al\mbox{.}(2016)]%
        {shou2016temporal}
\bibfield{author}{\bibinfo{person}{Zheng Shou}, \bibinfo{person}{Dongang Wang},
  {and} \bibinfo{person}{Shih-Fu Chang}.} \bibinfo{year}{2016}\natexlab{}.
\newblock \showarticletitle{Temporal action localization in untrimmed videos
  via multi-stage cnns}. In \bibinfo{booktitle}{\emph{Proceedings of the IEEE
  conference on computer vision and pattern recognition}}.
  \bibinfo{pages}{1049--1058}.
\newblock


\bibitem[Simonyan and Zisserman(2014)]%
        {simonyan2014two}
\bibfield{author}{\bibinfo{person}{Karen Simonyan} {and}
  \bibinfo{person}{Andrew Zisserman}.} \bibinfo{year}{2014}\natexlab{}.
\newblock \showarticletitle{Two-stream convolutional networks for action
  recognition in videos}.
\newblock \bibinfo{journal}{\emph{Advances in neural information processing
  systems}}  \bibinfo{volume}{27} (\bibinfo{year}{2014}).
\newblock


\bibitem[Soomro and Zamir(2014)]%
        {soomro2014action}
\bibfield{author}{\bibinfo{person}{Khurram Soomro} {and}
  \bibinfo{person}{Amir~R Zamir}.} \bibinfo{year}{2014}\natexlab{}.
\newblock \showarticletitle{Action recognition in realistic sports videos}.
\newblock In \bibinfo{booktitle}{\emph{Computer vision in sports}}.
  \bibinfo{publisher}{Springer}, \bibinfo{pages}{181--208}.
\newblock


\bibitem[Sri-Iesaranusorn et~al\mbox{.}(2021)]%
        {sri2021toward}
\bibfield{author}{\bibinfo{person}{Panyawut Sri-Iesaranusorn},
  \bibinfo{person}{Felan~Carlo Garcia}, \bibinfo{person}{Francis Tiausas},
  \bibinfo{person}{Supatsara Wattanakriengkrai}, \bibinfo{person}{Kazushi
  Ikeda}, {and} \bibinfo{person}{Junichiro Yoshimoto}.}
  \bibinfo{year}{2021}\natexlab{}.
\newblock \showarticletitle{Toward the Perfect Stroke: A Multimodal Approach
  for Table Tennis Stroke Evaluation}. In \bibinfo{booktitle}{\emph{2021
  Thirteenth International Conference on Mobile Computing and Ubiquitous
  Network (ICMU)}}. IEEE, \bibinfo{pages}{1--5}.
\newblock


\bibitem[Su et~al\mbox{.}(2021)]%
        {su2020bsn++}
\bibfield{author}{\bibinfo{person}{Haisheng Su}, \bibinfo{person}{Weihao Gan},
  \bibinfo{person}{Wei Wu}, \bibinfo{person}{Yu Qiao}, {and}
  \bibinfo{person}{Junjie Yan}.} \bibinfo{year}{2021}\natexlab{}.
\newblock \showarticletitle{Bsn++: Complementary boundary regressor with
  scale-balanced relation modeling for temporal action proposal generation}. In
  \bibinfo{booktitle}{\emph{Proceedings of the AAAI conference on artificial
  intelligence}}, Vol.~\bibinfo{volume}{35}. \bibinfo{pages}{2602--2610}.
\newblock


\bibitem[Thilakarathne et~al\mbox{.}(2022)]%
        {thilakarathne2021pose}
\bibfield{author}{\bibinfo{person}{Haritha Thilakarathne},
  \bibinfo{person}{Aiden Nibali}, \bibinfo{person}{Zhen He}, {and}
  \bibinfo{person}{Stuart Morgan}.} \bibinfo{year}{2022}\natexlab{}.
\newblock \showarticletitle{Pose is all you need: The pose only group activity
  recognition system (pogars)}.
\newblock \bibinfo{journal}{\emph{Machine Vision and Applications}}
  \bibinfo{volume}{33}, \bibinfo{number}{6} (\bibinfo{year}{2022}),
  \bibinfo{pages}{95}.
\newblock


\bibitem[Tran et~al\mbox{.}(2015)]%
        {tran2015learning}
\bibfield{author}{\bibinfo{person}{Du Tran}, \bibinfo{person}{Lubomir Bourdev},
  \bibinfo{person}{Rob Fergus}, \bibinfo{person}{Lorenzo Torresani}, {and}
  \bibinfo{person}{Manohar Paluri}.} \bibinfo{year}{2015}\natexlab{}.
\newblock \showarticletitle{Learning spatiotemporal features with 3d
  convolutional networks}. In \bibinfo{booktitle}{\emph{Proceedings of the IEEE
  international conference on computer vision}}. \bibinfo{pages}{4489--4497}.
\newblock


\bibitem[Voeikov et~al\mbox{.}(2020)]%
        {voeikov2020ttnet}
\bibfield{author}{\bibinfo{person}{Roman Voeikov}, \bibinfo{person}{Nikolay
  Falaleev}, {and} \bibinfo{person}{Ruslan Baikulov}.}
  \bibinfo{year}{2020}\natexlab{}.
\newblock \showarticletitle{TTNet: Real-time temporal and spatial video
  analysis of table tennis}. In \bibinfo{booktitle}{\emph{Proceedings of the
  IEEE/CVF Conference on Computer Vision and Pattern Recognition Workshops}}.
  \bibinfo{pages}{884--885}.
\newblock


\bibitem[Wadsworth et~al\mbox{.}(2020)]%
        {wadsworth2020use}
\bibfield{author}{\bibinfo{person}{Nick Wadsworth}, \bibinfo{person}{Lewis
  Charnock}, \bibinfo{person}{Jamie Russell}, {and} \bibinfo{person}{Martin
  Littlewood}.} \bibinfo{year}{2020}\natexlab{}.
\newblock \showarticletitle{Use of video-analysis feedback within a six-month
  coach education program at a professional football club}.
\newblock \bibinfo{journal}{\emph{Journal of Sport Psychology in Action}}
  \bibinfo{volume}{11}, \bibinfo{number}{2} (\bibinfo{year}{2020}),
  \bibinfo{pages}{73--91}.
\newblock


\bibitem[Wang et~al\mbox{.}(2019b)]%
        {wang2019football}
\bibfield{author}{\bibinfo{person}{Bin Wang}, \bibinfo{person}{Wei Shen},
  \bibinfo{person}{FanSheng Chen}, {and} \bibinfo{person}{Dan Zeng}.}
  \bibinfo{year}{2019}\natexlab{b}.
\newblock \showarticletitle{Football match intelligent editing system based on
  deep learning}.
\newblock \bibinfo{journal}{\emph{KSII Transactions on Internet and Information
  Systems (TIIS)}} \bibinfo{volume}{13}, \bibinfo{number}{10}
  (\bibinfo{year}{2019}), \bibinfo{pages}{5130--5143}.
\newblock


\bibitem[Wang et~al\mbox{.}(2023)]%
        {wang2023videomae}
\bibfield{author}{\bibinfo{person}{Limin Wang}, \bibinfo{person}{Bingkun
  Huang}, \bibinfo{person}{Zhiyu Zhao}, \bibinfo{person}{Zhan Tong},
  \bibinfo{person}{Yinan He}, \bibinfo{person}{Yi Wang}, \bibinfo{person}{Yali
  Wang}, {and} \bibinfo{person}{Yu Qiao}.} \bibinfo{year}{2023}\natexlab{}.
\newblock \showarticletitle{Videomae v2: Scaling video masked autoencoders with
  dual masking}. In \bibinfo{booktitle}{\emph{Proceedings of the IEEE/CVF
  Conference on Computer Vision and Pattern Recognition}}.
  \bibinfo{pages}{14549--14560}.
\newblock


\bibitem[Wang et~al\mbox{.}(2014)]%
        {wang2014action}
\bibfield{author}{\bibinfo{person}{Limin Wang}, \bibinfo{person}{Yu Qiao},
  \bibinfo{person}{Xiaoou Tang}, {et~al\mbox{.}}}
  \bibinfo{year}{2014}\natexlab{}.
\newblock \showarticletitle{Action recognition and detection by combining
  motion and appearance features}.
\newblock \bibinfo{journal}{\emph{THUMOS14 Action Recognition Challenge}}
  \bibinfo{volume}{1}, \bibinfo{number}{2} (\bibinfo{year}{2014}),
  \bibinfo{pages}{2}.
\newblock


\bibitem[Wang et~al\mbox{.}(2016)]%
        {wang2016temporal}
\bibfield{author}{\bibinfo{person}{Limin Wang}, \bibinfo{person}{Yuanjun
  Xiong}, \bibinfo{person}{Zhe Wang}, \bibinfo{person}{Yu Qiao},
  \bibinfo{person}{Dahua Lin}, \bibinfo{person}{Xiaoou Tang}, {and}
  \bibinfo{person}{Luc~Van Gool}.} \bibinfo{year}{2016}\natexlab{}.
\newblock \showarticletitle{Temporal segment networks: Towards good practices
  for deep action recognition}. In \bibinfo{booktitle}{\emph{European
  conference on computer vision}}. Springer, \bibinfo{pages}{20--36}.
\newblock


\bibitem[Wang et~al\mbox{.}(2019a)]%
        {wang2019knowledge}
\bibfield{author}{\bibinfo{person}{Shangfei Wang}, \bibinfo{person}{Longfei
  Hao}, {and} \bibinfo{person}{Qiang Ji}.} \bibinfo{year}{2019}\natexlab{a}.
\newblock \showarticletitle{Knowledge-augmented multimodal deep regression
  bayesian networks for emotion video tagging}.
\newblock \bibinfo{journal}{\emph{IEEE Transactions on Multimedia}}
  \bibinfo{volume}{22}, \bibinfo{number}{4} (\bibinfo{year}{2019}),
  \bibinfo{pages}{1084--1097}.
\newblock


\bibitem[Wang et~al\mbox{.}(2018)]%
        {wang2018non}
\bibfield{author}{\bibinfo{person}{Xiaolong Wang}, \bibinfo{person}{Ross
  Girshick}, \bibinfo{person}{Abhinav Gupta}, {and} \bibinfo{person}{Kaiming
  He}.} \bibinfo{year}{2018}\natexlab{}.
\newblock \showarticletitle{Non-local neural networks}. In
  \bibinfo{booktitle}{\emph{Proceedings of the IEEE conference on computer
  vision and pattern recognition}}. \bibinfo{pages}{7794--7803}.
\newblock


\bibitem[Wang et~al\mbox{.}(2021)]%
        {wang2021self}
\bibfield{author}{\bibinfo{person}{Xiang Wang}, \bibinfo{person}{Shiwei Zhang},
  \bibinfo{person}{Zhiwu Qing}, \bibinfo{person}{Yuanjie Shao},
  \bibinfo{person}{Changxin Gao}, {and} \bibinfo{person}{Nong Sang}.}
  \bibinfo{year}{2021}\natexlab{}.
\newblock \showarticletitle{Self-supervised learning for semi-supervised
  temporal action proposal}. In \bibinfo{booktitle}{\emph{Proceedings of the
  IEEE/CVF Conference on Computer Vision and Pattern Recognition}}.
  \bibinfo{pages}{1905--1914}.
\newblock


\bibitem[Wei et~al\mbox{.}(2022)]%
        {wei2021masked}
\bibfield{author}{\bibinfo{person}{Chen Wei}, \bibinfo{person}{Haoqi Fan},
  \bibinfo{person}{Saining Xie}, \bibinfo{person}{Chao-Yuan Wu},
  \bibinfo{person}{Alan Yuille}, {and} \bibinfo{person}{Christoph
  Feichtenhofer}.} \bibinfo{year}{2022}\natexlab{}.
\newblock \showarticletitle{Masked feature prediction for self-supervised
  visual pre-training}. In \bibinfo{booktitle}{\emph{Proceedings of the
  IEEE/CVF Conference on Computer Vision and Pattern Recognition}}.
  \bibinfo{pages}{14668--14678}.
\newblock


\bibitem[Wu et~al\mbox{.}(2020)]%
        {wu2020multigrid}
\bibfield{author}{\bibinfo{person}{Chao-Yuan Wu}, \bibinfo{person}{Ross
  Girshick}, \bibinfo{person}{Kaiming He}, \bibinfo{person}{Christoph
  Feichtenhofer}, {and} \bibinfo{person}{Philipp Krahenbuhl}.}
  \bibinfo{year}{2020}\natexlab{}.
\newblock \showarticletitle{A multigrid method for efficiently training video
  models}. In \bibinfo{booktitle}{\emph{Proceedings of the IEEE/CVF Conference
  on Computer Vision and Pattern Recognition}}. \bibinfo{pages}{153--162}.
\newblock


\bibitem[Wu et~al\mbox{.}(2023)]%
        {wu2022survey}
\bibfield{author}{\bibinfo{person}{Fei Wu}, \bibinfo{person}{Qingzhong Wang},
  \bibinfo{person}{Jiang Bian}, \bibinfo{person}{Ning Ding},
  \bibinfo{person}{Feixiang Lu}, \bibinfo{person}{Jun Cheng},
  \bibinfo{person}{Dejing Dou}, {and} \bibinfo{person}{Haoyi Xiong}.}
  \bibinfo{year}{2023}\natexlab{}.
\newblock \showarticletitle{A Survey on Video Action Recognition in Sports:
  Datasets, Methods and Applications}.
\newblock \bibinfo{journal}{\emph{IEEE Transactions on Multimedia}}
  \bibinfo{volume}{25} (\bibinfo{year}{2023}), \bibinfo{pages}{7943--7966}.
\newblock
\urldef\tempurl%
\url{https://doi.org/10.1109/TMM.2022.3232034}
\showDOI{\tempurl}


\bibitem[Xie et~al\mbox{.}(2018)]%
        {xie2018rethinking}
\bibfield{author}{\bibinfo{person}{Saining Xie}, \bibinfo{person}{Chen Sun},
  \bibinfo{person}{Jonathan Huang}, \bibinfo{person}{Zhuowen Tu}, {and}
  \bibinfo{person}{Kevin Murphy}.} \bibinfo{year}{2018}\natexlab{}.
\newblock \showarticletitle{Rethinking spatiotemporal feature learning:
  Speed-accuracy trade-offs in video classification}. In
  \bibinfo{booktitle}{\emph{Proceedings of the European conference on computer
  vision (ECCV)}}. \bibinfo{pages}{305--321}.
\newblock


\bibitem[Xu et~al\mbox{.}(2021)]%
        {xu2021boundary}
\bibfield{author}{\bibinfo{person}{Mengmeng Xu}, \bibinfo{person}{Juan-Manuel
  P{\'e}rez-R{\'u}a}, \bibinfo{person}{Victor Escorcia}, \bibinfo{person}{Brais
  Martinez}, \bibinfo{person}{Xiatian Zhu}, \bibinfo{person}{Li Zhang},
  \bibinfo{person}{Bernard Ghanem}, {and} \bibinfo{person}{Tao Xiang}.}
  \bibinfo{year}{2021}\natexlab{}.
\newblock \showarticletitle{Boundary-sensitive pre-training for temporal
  localization in videos}. In \bibinfo{booktitle}{\emph{Proceedings of the
  IEEE/CVF International Conference on Computer Vision}}.
  \bibinfo{pages}{7220--7230}.
\newblock


\bibitem[Yan et~al\mbox{.}(2019)]%
        {yan2019multi}
\bibfield{author}{\bibinfo{person}{Huan Yan}, \bibinfo{person}{Chunfeng Yang},
  \bibinfo{person}{Donghan Yu}, \bibinfo{person}{Yong Li},
  \bibinfo{person}{Depeng Jin}, {and} \bibinfo{person}{Dah~Ming Chiu}.}
  \bibinfo{year}{2019}\natexlab{}.
\newblock \showarticletitle{Multi-site user behavior modeling and its
  application in video recommendation}.
\newblock \bibinfo{journal}{\emph{IEEE Transactions on Knowledge and Data
  Engineering}} \bibinfo{volume}{33}, \bibinfo{number}{1}
  (\bibinfo{year}{2019}), \bibinfo{pages}{180--193}.
\newblock


\bibitem[Yeung et~al\mbox{.}(2016)]%
        {yeung2016end}
\bibfield{author}{\bibinfo{person}{Serena Yeung}, \bibinfo{person}{Olga
  Russakovsky}, \bibinfo{person}{Greg Mori}, {and} \bibinfo{person}{Li
  Fei-Fei}.} \bibinfo{year}{2016}\natexlab{}.
\newblock \showarticletitle{End-to-end learning of action detection from frame
  glimpses in videos}. In \bibinfo{booktitle}{\emph{Proceedings of the IEEE
  conference on computer vision and pattern recognition}}.
  \bibinfo{pages}{2678--2687}.
\newblock


\bibitem[Yuan et~al\mbox{.}(2016)]%
        {yuan2016temporal}
\bibfield{author}{\bibinfo{person}{Jun Yuan}, \bibinfo{person}{Bingbing Ni},
  \bibinfo{person}{Xiaokang Yang}, {and} \bibinfo{person}{Ashraf~A Kassim}.}
  \bibinfo{year}{2016}\natexlab{}.
\newblock \showarticletitle{Temporal action localization with pyramid of score
  distribution features}. In \bibinfo{booktitle}{\emph{Proceedings of the IEEE
  Conference on Computer Vision and Pattern Recognition}}.
  \bibinfo{pages}{3093--3102}.
\newblock


\bibitem[Yue-Hei~Ng et~al\mbox{.}(2015)]%
        {yue2015beyond}
\bibfield{author}{\bibinfo{person}{Joe Yue-Hei~Ng}, \bibinfo{person}{Matthew
  Hausknecht}, \bibinfo{person}{Sudheendra Vijayanarasimhan},
  \bibinfo{person}{Oriol Vinyals}, \bibinfo{person}{Rajat Monga}, {and}
  \bibinfo{person}{George Toderici}.} \bibinfo{year}{2015}\natexlab{}.
\newblock \showarticletitle{Beyond short snippets: Deep networks for video
  classification}. In \bibinfo{booktitle}{\emph{Proceedings of the IEEE
  conference on computer vision and pattern recognition}}.
  \bibinfo{pages}{4694--4702}.
\newblock


\bibitem[Zalluhoglu and Ikizler-Cinbis(2020)]%
        {zalluhoglu2020collective}
\bibfield{author}{\bibinfo{person}{Cemil Zalluhoglu} {and}
  \bibinfo{person}{Nazli Ikizler-Cinbis}.} \bibinfo{year}{2020}\natexlab{}.
\newblock \showarticletitle{Collective Sports: A multi-task dataset for
  collective activity recognition}.
\newblock \bibinfo{journal}{\emph{Image and Vision Computing}}
  \bibinfo{volume}{94} (\bibinfo{year}{2020}), \bibinfo{pages}{103870}.
\newblock


\bibitem[Zhu et~al\mbox{.}(2020)]%
        {zhu2020comprehensive}
\bibfield{author}{\bibinfo{person}{Yi Zhu}, \bibinfo{person}{Xinyu Li},
  \bibinfo{person}{Chunhui Liu}, \bibinfo{person}{Mohammadreza Zolfaghari},
  \bibinfo{person}{Yuanjun Xiong}, \bibinfo{person}{Chongruo Wu},
  \bibinfo{person}{Zhi Zhang}, \bibinfo{person}{Joseph Tighe},
  \bibinfo{person}{R Manmatha}, {and} \bibinfo{person}{Mu Li}.}
  \bibinfo{year}{2020}\natexlab{}.
\newblock \showarticletitle{A comprehensive study of deep video action
  recognition}.
\newblock \bibinfo{journal}{\emph{arXiv preprint arXiv:2012.06567}}
  (\bibinfo{year}{2020}).
\newblock


\end{thebibliography}

\appendix

\end{document}